\PassOptionsToPackage{table}{xcolor}
\documentclass{fairmeta_gr2}
\usepackage{amsmath,amssymb}
\usepackage{tabularx,array}
\usepackage{float}
\usepackage{wrapfig}
\usepackage{needspace}
\usepackage{xurl}
\let\BibliographyUrlBreaks\UrlBreaks
\useOriginalUrlSetting
\definecolor{TableHeader}{HTML}{C9DCEE}
\definecolor{TableStripe}{HTML}{EFF5FA}
\definecolor{PaperBlue}{HTML}{245C99}
\newcommand{\task}[1]{\textnormal{\textsc{#1}}}
\newcommand{\state}{W}
\newcommand{\domain}{\mathcal{X}}
\newcommand{\emetric}{\mathrm{EM}}
\newcommand{\tablehead}{\rowcolor{TableHeader}}
\newcommand{\FinishWrap}{\par
  \ifnum\value{WF@wrappedlines}>1
    \vspace{\dimexpr\value{WF@wrappedlines}\baselineskip\relax}
  \fi
  \WFclear}

\fancypagestyle{plain}{\fancyhf{}\renewcommand{\headrulewidth}{0pt}\fancyfoot[C]{\small\thepage}}
\title{{\setstretch{1}\fontsize{24}{28}\selectfont Grid-Bench:\par}
  \vspace{3pt}
  {\setstretch{1}\fontsize{16}{20}\selectfont
  Can LLMs Predict How Discrete Worlds Change?\par}}
\author[1]{Zixiang Xu}
\author[2]{Zirui Song}
\author[1]{Wanrong Zheng}
\author[3]{Chiyu Zhang}
\author[4]{Yue Huang}
\author[5]{Yuyang Peng}
\author[1]{Kexin Zhang}
\author[1]{Jinkun Chen}
\author[2]{Xiuying Chen}
\author[4]{Xiangliang Zhang}
\author[1]{Yan Liu}
\author[1]{Laurent Itti}
\author[3]{Frank Shyu}
\author[3]{Xiaoqiang Luo}
\author[3]{Sandeep Pandey}
\author[3]{Xi Liu}
\author[3]{Luke Simon}
\author[6]{Xiyang Hu}
\author[1]{Yue Zhao}
\renewcommand{\affiliationlist}{%
\affiliationformat[1]{University of Southern California}\par%
\affiliationformat[2]{Mohamed bin Zayed University of Artificial Intelligence}\par%
\affiliationformat[3]{Meta}\quad \affiliationformat[4]{University of Notre Dame}\par%
\affiliationformat[5]{University of California, Los Angeles}\quad \affiliationformat[6]{Arizona State University}\par%
}
\hypersetup{pdfauthor={Zixiang Xu, Zirui Song, Wanrong Zheng, Chiyu Zhang, Yue Huang, Yuyang Peng, Kexin Zhang, Jinkun Chen, Xiuying Chen, Xiangliang Zhang, Yan Liu, Laurent Itti, Frank Shyu, Xiaoqiang Luo, Sandeep Pandey, Xi Liu, Luke Simon, Xiyang Hu, Yue Zhao}}

\abstract{Spatial instructions describe how to move objects, transform regions, or modify
their surroundings. When several operations are applied in sequence, each acts
on a configuration shaped by earlier changes. How accurately can large language
models (LLMs) predict the resulting spatial state? We introduce Grid-Bench, an executable
benchmark in which eight families of operations transform discrete worlds
represented as two-, three-, and four-dimensional grids. Given an initial grid,
explicit rules, and an event history expressed as
text, a model predicts the complete final grid. Reference execution makes every
predicted value checkable, supporting comparisons across spatial operations,
history lengths, and workloads. Evaluating six LLMs shows that
complete-state accuracy falls sharply as histories grow, with substantial
divergence among models that perform similarly on single events. Models with
similar exact-match accuracy can nevertheless differ substantially in the number
of incorrect cells on shared instances. Even over long histories, some models
recover many correct changes beyond what copying the initial state would yield.
Grid-Bench makes both complete recovery and these differences in returned states
directly measurable across a range of spatial conditions.
}
\metadata[Website]{\url{https://grid-bench.github.io/}}
\metadata[GitHub]{\url{https://github.com/grid-bench/Grid-Bench}}
\metadata[Hugging Face]{\url{https://huggingface.co/datasets/xzx34/Grid-Bench}}
\hypersetup{pdftitle={Grid-Bench: Can LLMs Predict How Discrete Worlds Change?},
  pdfsubject={Spatial state prediction in discrete worlds}}

\begin{document}
\maketitle
\begingroup
\clubpenalties 1 10000
\section{Introduction}
\label{sec:introduction}
Spatial reasoning supports understanding scene descriptions and interpreting
instructions about objects and their surroundings
\citep{shi2022stepgame,long2016scone}. Such instructions appear in tasks ranging
from object manipulation to geometric design, where language specifies actions
or construction operations that change a spatial configuration
\citep{shridhar2021alfworld,khan2024text2cad}. When instructions describe a
sequence of operations, earlier changes can alter the spatial conditions on which
later operations depend. Moving an object, for example, can free one position
while blocking another, changing the outcome of a subsequent move. Predicting
the resulting configuration from text therefore involves relating the described
operations to the spatial states they successively produce. We investigate how
accurately large language models predict these outcomes across varied operations,
event histories, and spatial structures.

Existing benchmarks evaluate selected spatial relations and entity states
\citep{mirzaee2021spartqa,shi2022stepgame,kim2023entity}, while recent studies
also examine complete-grid and chess-state prediction from event sequences
\citep{chapra2026complexity,walker2026chess}. We introduce Grid-Bench to study
complete-state prediction in discrete spatial worlds, across operations that
draw on different spatial information. A reflection rearranges values within a specified geometric region;
region growth acts on a connected component whose boundary changes with the
state. The grid makes coordinates and adjacency explicit, providing a common
representation and output format for comparing predictions across these
operations.

\begin{figure}[!htbp]
\centering
\includegraphics[trim={2.75bp 1.35bp 3.1bp 0.25bp},clip,width=\linewidth]{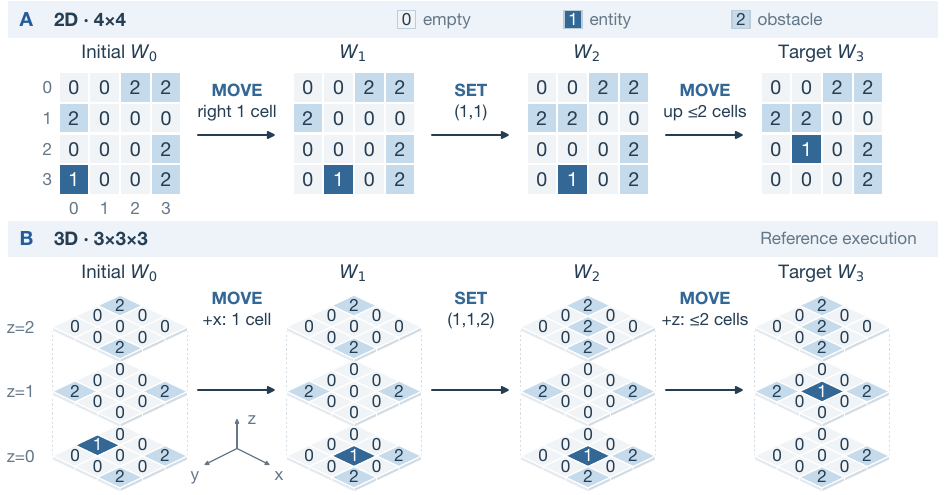}
\caption{Reference execution in 2D (A) and 3D (B), from $W_0$ through
$W_1$, $W_2$, and $W_3$. Values 0, 1, and 2 denote empty cells, the moving
entity, and obstacles. \task{MOVE} stops at an obstacle or its step limit;
\task{SET} places an obstacle at the labeled coordinate. The new obstacle
limits the final move to one cell. Coordinates are zero-based:
(row, column) in (A) and $(x,y,z)$ in (B). The model receives $W_0$, rules,
and all three events as text and predicts the complete $W_3$.}
\label{fig:overview}
\end{figure}

The benchmark covers eight operation families on two-, three-, and
four-dimensional grids, with instances varying in history length and
within-family workload. Each instance provides an initial grid, explicit rules,
and an ordered event sequence as text; the model predicts the complete final grid
after all supplied events in a single response. The rules specify how each event
selects locations and changes their values in the current configuration.
A simulator executes the full event sequence to compute an exact reference
state. Comparing every predicted value with this
reference establishes whether the whole state is correct and reveals the extent
and location of errors when it is not. Figure~\ref{fig:overview} illustrates
the reference states produced by a movement history in two and three
dimensions.

We evaluate six large language models under a shared protocol and find that
exact-match accuracy declines sharply with history length. Models with similar
single-event accuracy separate over longer histories, and even similar rates
of complete recovery can accompany quite different incorrect predictions.
At eight events, Kimi K2.5 and MiniMax M2.5 have similar exact-match accuracy,
yet their median wrong-cell counts are 14 and 67 on the 84 shared instances
where both return valid incorrect states.

Grid-Bench brings diverse spatial operations into a common, executable framework
for predicting and evaluating complete worlds. The accompanying six-model study
examines when models recover the final state across event histories and
spatial conditions, and what they get wrong when recovery fails.

\endgroup
\section{Related Work}
\label{sec:related}
Spatial benchmarks evaluate relations, transformations, and actions through
different prediction targets.
SPARTQA \citep{mirzaee2021spartqa} and SPARTUN/ReSQ
\citep{mirzaee2022transfer} ask questions about relations expressed in scene
descriptions. StepGame \citep{shi2022stepgame} tests relational composition,
while SpaRC/SpaRP \citep{rizvi2024sparc} formalizes spatial properties and
reasoning paths.
Other evaluations consider constrained path construction
\citep{kaesberg2025sparc}, spatial reasoning across textual and visual inputs
\citep{wang2024spatialeval}, and action selection in symbolic or simulated
environments \citep{gao2026cubebench,gao2026spatialworld}.
JigShape \citep{li2026jigshape} evaluates geometric reconstruction from puzzle
pieces, and OrigamiBench \citep{agarwal2026origami} studies physically grounded
fold operations.
SnorkelSpatial \citep{vishwakarma2025snorkelspatial} studies transformations of
textual boards, querying properties after translations and rotations.
Grid-Bench evaluates the complete configuration produced by a prescribed
event history, allowing errors to be located throughout the resulting world.

Predicting the result of an event history also connects spatial reasoning to
instruction execution and state tracking. bAbI \citep{weston2015babi} includes
questions about locations and possessions after narrated events; SCONE
\citep{long2016scone} grounds instruction sequences in executable worlds;
and ProPara \citep{dalvi2018propara} follows entity existence and location
through procedural text. Entity-tracking
evaluations more directly test whether models update affected entities while
retaining unaffected information across operations
\citep{kim2023entity,tang2026entities}. Beyond state tracking, GT Bench \citep{xu2026gta} examines
algorithmic reasoning over graphs, while AlgoWorlds \citep{xu2026algoworlds}
evaluates tool-guided decisions against verifiable optima.
Closest to our complete-state setting,
Alien Grid \citep{chapra2026complexity} studies sequential depth through
rotations, corner swaps, and shifts of a $3\times3$ grid. Chess-World-Model
\citep{walker2026chess} trains sequence models to recover complete chess states
at successive move prefixes. Grid-Bench evaluates LLMs given explicit textual
rules across geometric, neighborhood, connectivity, and movement operations,
using complete predictions to compare recovery across spatial conditions.
Appendix~\ref{app:related} develops these connections and compares the
corresponding evaluation settings.

\section{Grid-Bench}
\label{sec:benchmark}

\subsection{Spatial states and event histories}
\label{sec:state}
Grid-Bench asks for the complete spatial state after all events in a supplied
sequence.
A discrete world occupies a bounded $d$-dimensional grid with side lengths
$L_1,\ldots,L_d$ and domain
\[
\domain=\textstyle \prod_{j=1}^{d}\{0,\ldots,L_j-1\}.
\]
Cell values are integers from a finite alphabet $\mathcal{A}$ containing $0$. A state
$\state\in\mathcal{A}^{\domain}$ assigns one value to every coordinate.
Boundaries do not wrap, and $0$ denotes an empty cell.

The model receives the grid specification, initial state $\state_0$, explicit
rules, and the complete event sequence $h_T=(e_1,\ldots,e_T)$ as text.
It is asked to predict the state after all $T$ events in a single response.

Each event specifies an operation and its parameters. We use one core
transition family $f$ per history to measure recovery over repeated applications
of its operations. The sequence may also contain single-cell assignments
$\task{SET}(p,v)$, which write $v$ at $p$.
Let $\state_t$ denote the reference state after the first $t$ events.
The simulator applies the family's deterministic transition operator $F_f$:
\begin{equation}
\state_t=F_f(\state_{t-1},e_t),\qquad t=1,\ldots,T.
\label{eq:evolution}
\end{equation}
History length $T$ counts all supplied events, including these assignments;
operation workload is specified separately. At $T=0$, the target is the
initial state itself.

In Figure~\ref{fig:overview}, the three displayed events form $h_3$, and
Equation~\ref{eq:evolution} maps the displayed $\state_0$ to $\state_3$.
The \task{SET} between the two \task{MOVE} events places an obstacle that
limits the second move.

The requested prediction $\widehat{\state}_T\in\mathcal{A}^{\domain}$ uses the
same sparse format as the initial state: a list of all nonzero coordinate--value
pairs, with omitted coordinates taking value $0$.
The target is the entire reference world $\state_T$: every coordinate must
have its correct final value, including those whose values match the initial
state. Appendix~\ref{app:prompts} gives the model-facing prompts and worked
examples; Appendix~\ref{app:scoring} specifies the answer contract.

\subsection{Spatial transition families}
\label{sec:families}
The eight transition families in Table~\ref{tab:tasks} cover geometry,
neighborhood aggregation, connected regions, and movement. Explicit rules specify
how each event selects cells and transforms their values. Events may select a
region geometrically, as a ball or box, or through the current world's
connectivity, occupancy, and obstacles. These dependencies make an operation's
effect sensitive to the preceding history. Every family extends to all
supported dimensions under the shared state representation.

\begin{table}[!htbp]
\centering
\small
\begingroup
\renewcommand{\arraystretch}{1.08}
\rowcolors{2}{TableStripe}{white}
\begin{tabularx}{\textwidth}{lXX}
\toprule
\tablehead
\textbf{Rule} & \textbf{Spatial selection} & \textbf{State transformation} \\
\midrule
\task{PAINT} & Cells with a specified source value inside a Manhattan ball. & Simultaneously replace the source value with a target value. \\
\task{REFLECT} & An axis-aligned box. & Apply a bijective reflection along one axis, including zero-valued cells. \\
\task{STENCIL} & A Manhattan ball, with reads from adjacent cells beyond the ball. & Replace each selected value by a modular sum over its old local neighborhood. \\
\task{MIX} & Cells reachable within a bounded number of steps through nonzero cells. & Apply synchronous neighborhood aggregation while keeping updated cells nonzero. \\
\task{RELABEL} & The maximal equal-valued connected component containing a seed. & Assign one target value to the entire component. \\
\task{LAYER} & A component boundary or the component cells exposed to empty space. & Grow the component by one layer or erode its exposed layer. \\
\task{MOVE} & Entities of one label and the traversed positions on axis-parallel lines. & Move entities in a prescribed order until obstructed or the step limit is reached. \\
\task{COMPRESS} & Every line parallel to a selected axis. & Remove gaps, merge adjacent equal values once per pair, and pack toward one boundary. \\
\bottomrule
\end{tabularx}
\endgroup
\caption{The eight core transition families. Each trajectory uses core events
from one family, interleaved with single-cell \task{SET} assignments. Full definitions,
valid parameter ranges, and edge cases appear in Appendix~\ref{app:tasks}.}
\label{tab:tasks}
\end{table}

For example, a \task{LAYER} growth event fills the empty cells adjacent to a
selected component.
Let $N_1(x)=\{y\in\domain:\lVert x-y\rVert_1=1\}$ denote the in-bounds
axis-neighbors of $x$. For an occupied seed $p$, let $C=C_{\state}(p)$ be
its maximal axis-connected component of value $\state(p)$. The component's
external boundary and the region written by a growth event are
\begin{align*}
\partial C&=\{y\in\domain\setminus C:N_1(y)\cap C\ne\varnothing\},\\
S_{+}(\state,p)&=\{y\in\partial C:\state(y)=0\}.
\end{align*}
Growth fills these empty boundary cells with $\state(p)$; erosion instead
clears component cells with an in-bounds empty neighbor. The component and
boundary are computed before writing, so each event acts on a single layer.
Earlier events can enlarge or split the component, change its boundary, or
clear its seed. The same later instruction can therefore affect a different
region depending on the preceding history.

Read and write regions can differ: \task{STENCIL} updates a ball using
values from selected cells and their neighbors, including those outside the
ball. These updates read the same pre-event state; \task{MOVE} instead processes
entities front to back, exposing vacated positions to later entities within
the event.
Appendices~\ref{app:tasks} and~\ref{app:rule-properties} give full transition
equations, read/write accounting, and conservation and ordering properties.

\subsection{Reference trajectories and instance construction}
\label{sec:construction}
The construction varies history length, dimension, and within-family workload
to study prediction across different spatial conditions.
Dimension comparisons use approximately equal world volumes
$N=|\domain|=\prod_{j=1}^{d}L_j$, varying coordinate arity and neighborhood
structure while keeping the number of predicted cell values nearly constant.
Operation scale is specified by a family-specific workload target
$w$: cell-value reads during reference execution for most families,
update-region size for \task{STENCIL} and \task{MIX}, and world volume for
\task{COMPRESS}.
Realized workload, value changes, and core-event counts are recorded separately
from the target $w$ and history length $T$.

A generation condition fixes the transition family, grid shape, alphabet,
target workload, and policies for initial states and event proposals. Starting
from a sampled $\state_0$, the simulator executes candidate events and selects
among them using realized workload and resulting occupancy. Accepted trajectories
remain nonempty and satisfy bounds on density, mean workload, and the fraction
of core events that change the state. Settings were frozen before formal
evaluation;
Appendix~\ref{app:generation} gives the generation settings and the
selection and acceptance procedures.

Each accepted trajectory supplies instances at selected prefix lengths.
A length-$T$ request contains the original state $\state_0$ and the first $T$
events $(e_1,\ldots,e_T)$, with the target $\state_T$ computed by
Equation~\ref{eq:evolution}. These prefixes are queried separately, allowing
comparison across progressively longer histories from the same initial world.
The shared reference trace records intermediate states and event statistics
for exact replay.

\subsection{Evaluating complete-state predictions}
\label{sec:measurement}
Complete predictions support evaluation of both exact recovery and cell-level
agreement with the reference.
The primary measure is semantic full-world exact match (EM). For request $i$,
let $\state_i$ be the reference final state and $v_i$ indicate validity of
the extracted prediction $\widehat{\state}_i$. A prediction is valid if it
decodes to a grid satisfying the output specification. Set $y_i=1$ when the prediction
is valid and equals $\state_i$ at every coordinate, and $y_i=0$ otherwise.
For requests $I$,
\begin{equation}
\emetric(I)=\textstyle \frac{1}{|I|}\sum_{i\in I}y_i.
\label{eq:em}
\end{equation}
Semantic equality compares cell values, including implicit zeros, regardless
of list ordering. Invalid or absent states score zero in the full denominator.

Exact match treats all incorrect states alike; cell-level comparisons
distinguish a few wrong values from errors across the grid.
For a valid prediction on domain $\domain_i$, the error set
$E_i=\{x\in\domain_i:\widehat{\state}_i(x)\ne\state_i(x)\}$ gives the
wrong-cell count $|E_i|$. For the valid subset $V=\{i\in I:v_i=1\}$, cell
accuracy averages the fraction of correct cells in each returned world:
\begin{equation}
\operatorname{CA}(I)
=\textstyle \frac{1}{|V|}\sum_{i\in V}
\left(1-\frac{|E_i|}{|\domain_i|}\right).
\label{eq:cell-accuracy}
\end{equation}
This macro-average gives each valid prediction equal weight. Error-extent
summaries instead use valid incorrect predictions,
$V^{-}=\{i\in V:y_i=0\}$, and report the subset size, median wrong-cell count,
and 90th percentile.

We further distinguish errors on cells that must change from errors on cells
whose final values must match the initial world. With $\state_{0,i}$ denoting
the initial state, define
\begin{equation}
\begin{aligned}
C_i&=\{x\in\domain_i:\state_i(x)\ne\state_{0,i}(x)\},\\
U_i&=\domain_i\setminus C_i.
\end{aligned}
\label{eq:change-partition}
\end{equation}
This partition uses initial and final values; a cell in $U_i$ may change
during the history and later return to its original value. Each valid
incorrect prediction is classified as \emph{changed-only} if its errors lie
entirely in $C_i$, \emph{unchanged-only} if they lie entirely in $U_i$, or
\emph{both} if they intersect both sets. In Figure~\ref{fig:overview}, retaining 1 at
$(3,0)$ in (A) would be an error in $C_i$; clearing $(2,0,0)$ in (B) would
be an error in $U_i$. Appendix~\ref{app:scoring} gives
the extraction and validity rules, answer accounting, cell-group accuracies,
quantile convention, and empty-subset handling.

\section{Experiments and Results}
\label{sec:experiments}

\subsection{Experimental setup}
\label{sec:setup}
We evaluate GPT-5.6 Sol, Claude 5 Sonnet, GPT-5.4 Mini, GLM-5, Kimi K2.5,
and MiniMax M2.5 on a shared set of 1,488 instances, yielding 8,928 model
outcomes. All six models receive identical initial worlds, event histories,
and prompts specifying the rules and task instructions. Predictions are checked
against the same reference states. Table~\ref{tab:evaluation-coverage}
summarizes the evaluation coverage.

\begin{table}[!htbp]
\centering
\small
\begingroup
\setlength{\tabcolsep}{4pt}
\renewcommand{\arraystretch}{1.08}
\begin{tabularx}{\columnwidth}{X*{2}{>{\centering\arraybackslash}X}}
\toprule
\tablehead
Evaluation & \shortstack{Instances\\per model} & \shortstack{Six-model\\outcomes} \\
\midrule
\rowcolor{TableStripe}
Main & 1,152 & 6,912 \\
\rowcolor{white}
Additional workload & 336 & 2,016 \\
\rowcolor{TableStripe}
Unique total & 1,488 & 8,928 \\
\bottomrule
\end{tabularx}
\endgroup
\caption{Evaluation coverage. Lengths are $T=0,1,8,16,32,64$. The workload analysis combines seven additional settings with four reused main settings, for 528 instances per model; reused instances are counted only once in the total.}
\label{tab:evaluation-coverage}
\end{table}

The main experiment crosses eight task families with three spatial dimensions
and six history lengths, $T=0,1,8,16,32,64$. The grids contain 252 or 256
cells, allowing comparisons across spatial structures at approximately matched
world volume. A complementary workload study varies operation scale at $d=2$
for \task{PAINT}, \task{REFLECT}, \task{LAYER}, and \task{MIX}. Across both
studies, 248 independently generated trajectories supply the evaluated
instances: eight trajectories per generation condition, each queried separately
at six prefixes.

Each request supplies the full task as text, permits one response with a
128,000-token combined reasoning-and-answer ceiling, and provides no tools.
We request high reasoning effort where supported; MiniMax M2.5 uses native
always-on thinking. The full sample accounting and request parameters appear
in Appendix~\ref{app:collection}, with the shared prompt in
Appendix~\ref{app:prompts}.

We report full-world EM over all requests and analyze returned states using
the measures in Section~\ref{sec:measurement}. Pooled summaries use
$T=8,16,32,64$ with equal length weights. The length curves use marginal
95\% intervals from 2,000 stratified bootstrap resamples of whole trajectories,
keeping their nested prefixes together as specified in
Appendix~\ref{app:statistics}. Complete supplementary results and resource
records are indexed in Appendix~\ref{app:data-records}.

\Needspace{24\baselineskip}
\subsection{Complete-state recovery across event histories}
\label{sec:length}
\begin{wrapfigure}{r}{0.48\textwidth}
\vspace{-\intextsep}
\centering
\includegraphics[trim={5bp 13.51bp 4.8bp 6.75bp},clip,width=\linewidth]{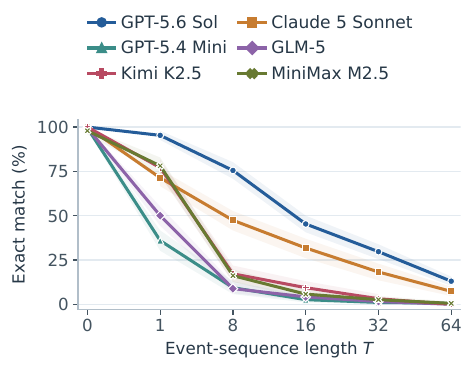}
\caption{Complete-state recovery across event histories. Each point reports
full-world EM on 192 main instances. Colors and markers identify models; shading gives marginal 95\%
trajectory-bootstrap intervals, omitting degenerate intervals. Lengths are categorical.}
\label{fig:length}
\end{wrapfigure}
At $T=0$, all six models recover the supplied initial state almost perfectly.
Exact-match accuracy then declines sharply as the same worlds undergo longer
event histories. In Figure~\ref{fig:length}, GPT-5.6 Sol's accuracy
falls from 95\% after one event to 13\% at $T=64$.

Longer histories also distinguish models with similar one-event accuracy.
Claude 5 Sonnet, Kimi K2.5, and MiniMax M2.5 begin with comparable scores,
yet the latter two fall well below Claude 5 Sonnet by $T=8$. The corresponding
paired and collection-cohort comparisons appear in
Appendix~\ref{app:paired-results}.

Within the overall decline, success remains uneven along individual
histories. Every model has cases in which a later, separately queried prefix
is correct after an earlier positive-length prefix was incorrect.
Appendix~\ref{app:prefix-patterns} documents these reversals on shared
reference histories.
\FinishWrap

\begin{figure}[!htbp]
\centering
\includegraphics[trim={10.25bp 6.3bp 6.1bp 10.25bp},clip,width=\linewidth]{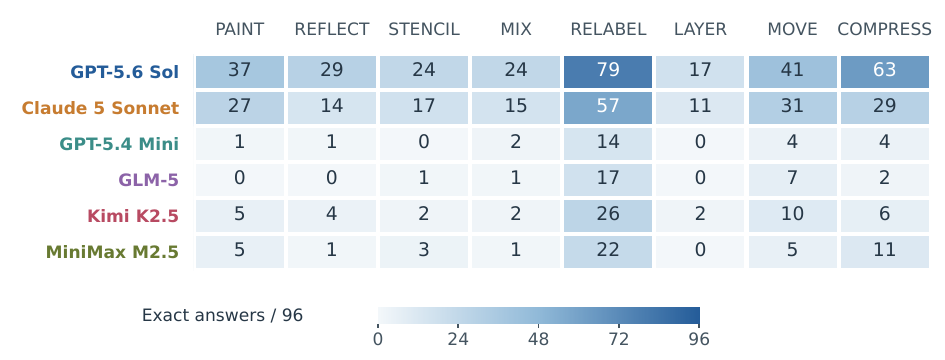}
\caption{Complete-state recovery across eight task families. Each cell gives
exact answers out of 96 instances, comprising four prefixes
($T=8,16,32,64$) of 24 trajectories across three dimensions.
Shading encodes counts on a shared scale; label colors identify models.}
\label{fig:tasks}
\end{figure}
\FloatBarrier

\subsection{Errors in predicted states}
\label{sec:state-errors}

Error extent and location are measured on valid incorrect grids;
cell accuracy and baseline comparisons also include exact grids.

\begin{figure}[!htbp]
\centering
\includegraphics[trim={9.25bp 1.35bp 9.6bp 2bp},clip,width=\linewidth]{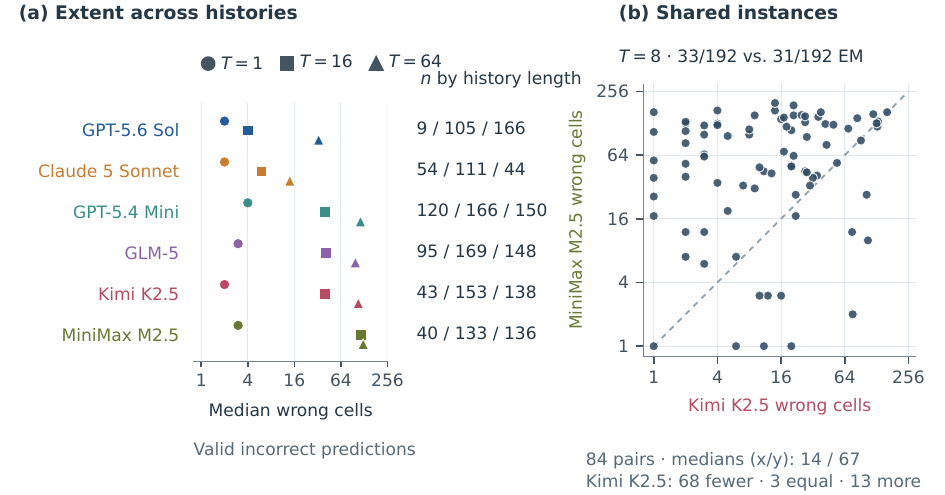}
\caption{Error extent in returned states.
(a) Median wrong-cell counts among each model's valid incorrect answers at
$T=1,16,64$ on a logarithmic axis. Marker shapes identify lengths, and the
$n$ column lists their sample sizes in that order. Each length uses its own
conditional subset.
(b) Wrong-cell counts on 84 shared instances where Kimi K2.5 and MiniMax M2.5
both return valid incorrect states at $T=8$. Both axes are logarithmic with equal scales; points above the
equality line indicate fewer errors for Kimi K2.5.}
\label{fig:error-contrast}
\end{figure}

Among these incorrect grids, one-event predictions
are typically close to the target, whereas every model has a larger median
wrong-cell count at the longest history in Figure~\ref{fig:error-contrast}(a).
GPT-5.6 Sol illustrates this change: most of its incorrect states at $T=16$
differ from the target in at most five cells, but such small deviations
become uncommon at $T=64$. The error tails in Appendix~\ref{app:error-extent}
show how far these departures extend beyond the typical prediction.

\begin{figure}[!htbp]
\centering
\includegraphics[trim={18bp 5.99bp 19.6bp 3.25bp},clip,width=\linewidth]{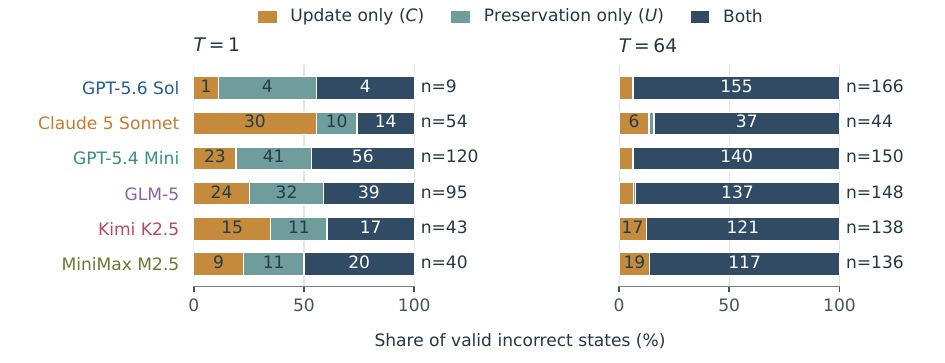}
\caption{Composition of valid incorrect states at $T=1$ and $T=64$.
Each bar shows the proportion with errors on reference-changed cells only
($C$), reference-unchanged cells only ($U$), or both. Segments display counts
where space permits; $n$ counts the conditional subset at each length.
Changed and unchanged compare final reference values with initial values.}
\label{fig:state-errors}
\end{figure}
\FloatBarrier

\Needspace{7\baselineskip}
This variation in error extent also separates models that appear similar
under exact match. Figure~\ref{fig:error-contrast}(b) illustrates this difference
for Kimi K2.5 and MiniMax M2.5 at $T=8$.
On the 84 shared instances with an incorrect grid from both models,
their median wrong-cell counts are 14 and 67, respectively. The contrast
appears in both collection cohorts and also when the shared comparison includes
exact answers, as detailed in Appendices~\ref{app:cohort-results}
and~\ref{app:matched-valid}.

\FloatBarrier

Long-history errors extend across both updating and preservation requirements.
The model-specific one-event patterns in Figure~\ref{fig:state-errors} give
way to a common pattern at $T=64$: most incorrect grids have errors
both on cells that should change and on cells whose final values should
match the initial world. At $T=16,32,64$, every model nevertheless has
higher cell accuracy on final-unchanged cells, as reported in
Appendix~\ref{app:change-diagnostics}.
Much of the information that should be preserved can therefore remain
correct even when a prediction contains errors in both groups.
Appendix~\ref{app:failure-cases} illustrates these patterns with actual predictions.

Some incomplete predictions recover correct changes well beyond what copying
the initial world would achieve. Both copying the initial state and returning
zeros match many individual values, although neither recovers a complete
positive-length main target.
Table~\ref{tab:revision-baselines} compares predictions and baselines on the
same instances within each row at $T=64$, with their coverage of all requests.
GPT-5.6 Sol exceeds copying by about 31 percentage points in mean
cell accuracy, whereas GPT-5.4 Mini and MiniMax M2.5 have much smaller gains.
The instance-level comparisons in Appendix~\ref{app:baseline-accuracy} show
that GPT-5.6 Sol improves on copying across almost all of its returned grids,
including many that fail exact match.

\begin{table}[!htbp]
\centering
\small
\begingroup
\setlength{\tabcolsep}{4pt}
\renewcommand{\arraystretch}{1.08}
\begin{tabularx}{\columnwidth}{lr*{3}{>{\centering\arraybackslash}X}}
\toprule
\tablehead
\textbf{Model} & $n/N$ & \textbf{Pred.} & \textbf{Copy} & \textbf{Zero} \\
\midrule
\rowcolor{TableStripe}
GPT-5.6 Sol & 191/192 & 81.6 & 51.0 & 48.8 \\
\rowcolor{white}
Claude 5 Sonnet & 58/192 & 90.5 & 49.5 & 56.5 \\
\rowcolor{TableStripe}
GPT-5.4 Mini & 151/192 & 56.1 & 50.6 & 49.8 \\
\rowcolor{white}
GLM-5 & 148/192 & 62.4 & 52.5 & 50.6 \\
\rowcolor{TableStripe}
Kimi K2.5 & 138/192 & 55.6 & 51.1 & 50.2 \\
\rowcolor{white}
MiniMax M2.5 & 137/192 & 52.4 & 50.9 & 49.4 \\
\bottomrule
\end{tabularx}
\endgroup
\caption{Cell accuracy (percent) at $T=64$. $n/N$ gives valid grids, including exact ones, out of all main requests. Pred. denotes the model prediction; Copy and Zero denote copy-initial and all-zero. All three use the same instances within each row; subsets differ across models.}
\label{tab:revision-baselines}
\end{table}

Executing only the later core operations rarely recovers the complete final
state. Starting from the initial world, a partial-execution baseline follows
the supplied event order, applying every \task{SET} assignment but executing core operations
only in the final half of the history. Only three of the 192 main targets at
$T=64$ are recovered this way.
Appendix~\ref{app:baseline-accuracy} reports the shorter suffixes.

\subsection{Performance across spatial conditions}
\label{sec:structure}
\label{sec:rules}
Low aggregate accuracy at long histories combines little or no recovery on
some task conditions with continued success on others. Figure~\ref{fig:tasks}
\mbox{summarizes} this concentration over $T=8$--64. \task{RELABEL} is the
highest-scoring family for every model under its low-merger policy, while
the relative performance across the remaining families varies by model.
The length curves in Appendix~\ref{app:results-context} identify where
recovery persists as histories grow; Appendix~\ref{app:relabel-distribution}
specifies the sampled connectivity conditions.

Approximately matching world volume also leaves substantial differences
across spatial structures. Pooled over $T=8$--64, accuracy is nonincreasing
from two through four dimensions for every model. The early profiles in
Table~\ref{tab:dimension-summary} differ: GLM-5 already has a pronounced
dimension difference after one event, whereas GPT-5.6 Sol has nearly equal
one-event scores. Its advantage in two dimensions emerges over longer
histories. The full profiles in Appendix~\ref{app:structural-context}
show the early reversals, while Appendix~\ref{app:geometry} describes the
corresponding geometry.

\begin{table}[!htbp]
\centering
\small
\begingroup
\setlength{\tabcolsep}{3pt}
\renewcommand{\arraystretch}{1.08}
\begin{tabularx}{\columnwidth}{l*{6}{>{\centering\arraybackslash}X}}
\toprule
\tablehead
 & \multicolumn{3}{c}{$T=1$} & \multicolumn{3}{c}{$T=16$} \\
\tablehead
\textbf{Model} & \mbox{$d{=}2$} & \mbox{$d{=}3$} & \mbox{$d{=}4$} & \mbox{$d{=}2$} & \mbox{$d{=}3$} & \mbox{$d{=}4$} \\
\midrule
\rowcolor{TableStripe}
GPT-5.6 Sol & 62 & 60 & 61 & 49 & 21 & 17 \\
\rowcolor{white}
Claude 5 Sonnet & 53 & 52 & 32 & 34 & 16 & 11 \\
\rowcolor{TableStripe}
GPT-5.4 Mini & 29 & 26 & 14 & 4 & 1 & 0 \\
\rowcolor{white}
GLM-5 & 47 & 30 & 19 & 6 & 2 & 0 \\
\rowcolor{TableStripe}
Kimi K2.5 & 56 & 53 & 39 & 13 & 5 & 0 \\
\rowcolor{white}
MiniMax M2.5 & 55 & 51 & 44 & 6 & 5 & 0 \\
\bottomrule
\end{tabularx}
\endgroup
\caption{Exact answers out of 64 main instances per cell, retaining all failures. The two lengths show early and intermediate structural contrasts. Worlds at $d=2,3,4$ contain 256, 252, and 256 cells; geometry and coordinate representation change together. Complete length curves appear in Appendix Figure~\ref{fig:structure}.}
\label{tab:dimension-summary}
\end{table}

\begin{figure}[!htbp]
\centering
\includegraphics[trim={18bp 9.8bp 5.85bp 6.25bp},clip,width=\linewidth]{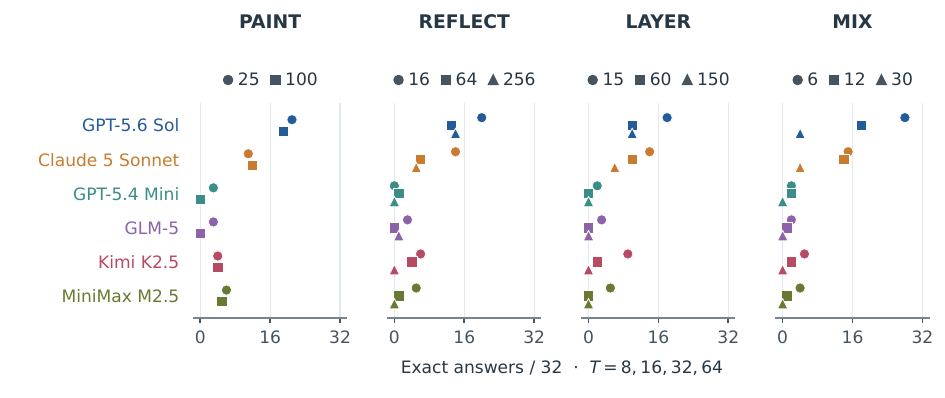}
\caption{Within-family workload comparisons at $d=2$. Each panel shows all
six models; marker shapes denote the task-specific targets listed above it.
Points report exact answers out of 32 instances: four prefixes
($T=8,16,32,64$) from eight trajectories per target. Targets refer to ball size
(\task{PAINT}), box volume (\task{REFLECT}), component-plus-boundary size
(\task{LAYER}), and reached-region size (\task{MIX}).}
\label{fig:workload-summary}
\end{figure}

Operation scale has a more task-dependent relationship with recovery.
In the pooled $T=8$--64 workload comparison at $d=2$ in
Figure~\ref{fig:workload-summary}, every model has lower accuracy at the
largest \task{MIX} and \task{LAYER} targets than at the smallest.
Comparing the smallest and largest \task{MIX} workload targets at $T=64$,
reference execution records roughly fivefold increases in both mean cell-value
reads and mean changed-cell counts per core event.
\task{PAINT} instead has mixed directions across models, and \task{REFLECT}
includes nonmonotonic profiles. Appendix~\ref{app:workload-profiles}
examines these length-specific \mbox{contrasts} and realized workloads.

\FloatBarrier

\section{Conclusion}
\label{sec:conclusion}
Grid-Bench uses eight families of executable operations and exact reference
states to evaluate how accurately language models predict the complete final
state of a discrete spatial world.
Across six models, success on individual operations gives way to much lower
exact recovery over longer histories, with substantial differences across
models and spatial conditions. When complete recovery fails, the returned
states range from nearly correct grids to extensively incorrect configurations.
Some retain many correct updates beyond what copying the initial state would
recover, even at the longest histories. Reliable prediction of spatial change
remains challenging even with explicit rules. The executable rules can supply
training examples with exact targets, while cell-level diagnostics can show
whether models improve at predicting required changes, preserving unchanged
values, or both.

\FloatBarrier
\section*{Limitations}
Grid-Bench evaluates predictions in fully specified symbolic worlds with
prescribed operations. Its task families and generation settings define the
spatial conditions studied here. Transfer to other world sizes, operation
distributions, or environments with uncertain observations remains an empirical
question.

Error extent and update--preservation partitions describe the final values
models return. Connecting these outcomes to internal representations or
execution strategies requires evidence about intermediate computation.

\section*{Acknowledgment}
All research and engineering work reported in this paper, including the design,
implementation, evaluation, and public release of Grid-Bench, was carried out by
the university-affiliated authors. The Meta-affiliated authors' contributions
were advisory only.

\begingroup
\setlength{\bibsep}{2.5pt plus 0.5pt}
\let\UrlBreaks\BibliographyUrlBreaks
\Urlmuskip=0mu plus 1mu
\bibliographystyle{assets/plainnat}
\bibliography{references,references-arxiv}
\endgroup
\clearpage
\appendix
\section{Extended Related Work}
\label{app:related}
Spatial state prediction connects spatial evaluation with tasks that map a
described world and a sequence of operations to an observable outcome. This
appendix develops connections to spatial evaluation, instruction execution
and entity state, complete-state prediction, structured reasoning, and
contextual robustness. Table~\ref{tab:related} compares representative
prediction targets and evaluation emphases.

\subsection{Spatial relations, perception, and action}
\label{app:related-spatial}
Textual spatial benchmarks have progressively broadened the relations and
compositions exposed in language. SPARTQA asks relation, object, and containment
questions over human-written and generated scene descriptions
\citep{mirzaee2021spartqa}; SPARTUN and ReSQ broaden relation types, linguistic
expressions, and human-authored questions \citep{mirzaee2022transfer}. StepGame
varies the length of relation chains and the presence of distractors
\citep{shi2022stepgame}, and later analysis separates relation parsing from
logical composition while correcting template artifacts
\citep{li2024advancing}. SpaRC formalizes properties of objects, relations, and
contexts, while SpaRP supplies deductively verified reasoning paths
\citep{rizvi2024sparc}.

Visual benchmarks provide a complementary tradition of controlled spatial
evaluation. CLEVR uses rendered scenes and generated questions to diagnose
compositional visual reasoning \citep{johnson2017clevr}; NLVR pairs natural
statements with synthetic images requiring visual and set-theoretic reasoning
\citep{suhr2017nlvr}. VSR evaluates natural text--image pairs across a broad
inventory of spatial relations and reference frames \citep{liu2023vsr}, while
SpatialEval compares spatial questions across textual and visual inputs
\citep{wang2024spatialeval}. These benchmarks make relation recognition and
composition observable under different balances of perception and language.
JigShape extends geometric evaluation to reconstructing jigsaw arrangements
from interlocking pieces and visual content \citep{li2026jigshape}. OrigamiBench
connects visual observations with a language of physically grounded fold
actions \citep{agarwal2026origami}. FigEdit studies another form of structured
transformation: checking whether chart edits respect the intended data and
graphical relationships \citep{li2026figedit}.

Maintaining a coherent visual world across time introduces further requirements.
CAT-NeRF separates appearance shared across a video from frame-specific body
variation \citep{zhu2023catnerf}, while Memorize-and-Generate compresses prior
frames into memory to retain historical scene information during video
generation \citep{zhu2025mag}. VISD uses feedback on answer correctness,
logical consistency, and spatiotemporal grounding to train video reasoning
models \citep{lin2026visd}. Learning a symbolic description from observations
is a related problem: abductive state grounding uses environmental knowledge
to supervise the mapping from sensory inputs to symbolic states
\citep{su2026abductive}.

Changes in the external world also affect the knowledge needed to answer visual
questions. LiveVQA evaluates seeking and updating recent visual knowledge
\citep{fu2025livevqa}; Worldwide LiveVQA extends this setting across languages,
with entity-recognition and multi-hop questions \citep{huang2026worldwide}.
These evaluations concern keeping knowledge current as new observations become
available. Grid-Bench supplies the initial world and the events that determine
the requested final state.

Other work places spatial reasoning inside path construction or interactive
action selection. SPaRC pathfinding asks models to construct grid paths under
interacting spatial, arithmetic, and geometric constraints
\citep{kaesberg2025sparc}. TextWorld generates executable text games with
controllable worlds and quests \citep{cote2018textworld}; ALFWorld aligns such
textual interaction with embodied household environments
\citep{shridhar2021alfworld}, and ScienceWorld provides an interactive text
environment for scientific tasks \citep{wang2022scienceworld}. More recent
spatial-agent evaluations include iVISPAR's sliding-tile tasks across text, 2D,
and 3D inputs \citep{mayer2025ivispar}, CubeBench's symbolic and visual cube
manipulation \citep{gao2026cubebench}, and SpatialWorld's egocentric multimodal
environments \citep{gao2026spatialworld}. Three-Step Nav combines landmark
planning, local visual guidance, and trajectory review for language-guided
navigation \citep{zheng2026threestep}. Ponder \& Press separates instruction
interpretation from locating interface elements for computer control
\citep{wang2025ponder}. These settings evaluate paths or
actions selected toward a goal. Grid-Bench instead supplies the operations and
evaluates the complete configuration they produce.

\subsection{Instruction execution and entity state}
\label{app:related-tracking}
Controlled language tasks have long connected narrated events to changing
worlds. bAbI generates questions about locations, possessions, and other facts
after sequences of events \citep{weston2015babi}. SCONE pairs an initial world
with context-dependent instructions in three executable domains; predicted
logical forms are evaluated by the world they produce
\citep{long2016scone}. Both use simulation to make consequences checkable, while
placing different demands on the predicted answer.

Procedural-text research studies changes that may be implicit in natural
descriptions. Neural Process Networks model actions as operators that transform
learned entity states \citep{bosselut2018process}. ProPara annotates entity
existence and location as a process unfolds \citep{dalvi2018propara}, and OpenPI
extends procedural state changes to open-vocabulary entities, attributes, and
before/after values \citep{tandon2020openpi}. TRIP associates multi-sentence
physical stories with state annotations that support tiered evaluation of
plausibility and its underlying evidence \citep{storks2021trip}. This line asks
models to infer consequences from ordinary language; Grid-Bench makes the
transition rules explicit and varies the spatial structures on which they act.

Entity-tracking studies isolate update and retention more directly. Box-transfer
tasks query container contents after PUT, REMOVE, and MOVE operations and
distinguish changed from unchanged states \citep{kim2023entity}. Subsequent work
uses behavioral tests, probes, and interventions to study how models process
these changes \citep{tang2026entities}. Related probing work recovers and
manipulates entity-state information in models trained on Alchemy instructions
and TextWorld transcripts \citep{li2021implicit}. These studies motivate
examining both final values that differ from the initial world and those that
match it when a model returns an entire state.

The use of evolving histories also appears in social and personalized
interaction. SocialMaze evaluates reasoning across social settings with
changing interactions and uncertain information \citep{xu2025socialmaze};
PersonaConvBench evaluates personalized predictions and generation from
conversation histories \citep{li2025personaconvbench}. Memory Retrieval for
Changing Preferences selects historical turns as evidence of a user's current
preferences \citep{qin2026memory}, while RaMem preserves temporal and
participant context when retrieving past experiences \citep{yang2026ramem}.
The scope of retained information matters as well: shared-state agents can
misapply information from one user's interaction to another
\citep{yang2026contamination}. For procedural memory, HyperSkill links reusable
skills and subtask steps through the trajectories in which they occurred
\citep{xu2026hyperskill}. This literature examines which past information
remains useful and how it should be represented; spatial-state prediction
provides explicit operations whose consequences can be checked at the cell level.

\subsection{Complete-state prediction}
\label{app:related-full-state}
Board games provide deterministic state transitions over long sequences.
Language modeling of chess move records has been evaluated with piece-location
and legal-move probes \citep{toshniwal2022chess}; Othello-GPT work uses probes
and interventions to study board representations learned from move transcripts
\citep{li2023othello}. These studies connect sequence modeling to an underlying
board state through targeted queries or internal analyses.

Two recent benchmarks make complete state recovery the evaluated output.
Chess-World-Model predicts 75 categorical labels for board occupancy and
auxiliary chess state at successive move prefixes, reporting both exact and
labelwise accuracy \citep{walker2026chess}. The Alien Grid domain of the
Complexity Ceiling Benchmark requests intermediate and final $3\times3$ grids
after rotations, corner swaps, and shifts at increasing sequential depths
\citep{chapra2026complexity}. SnorkelSpatial likewise applies translations and
rotations to textual boards and particles, then asks targeted questions about
the transformed result \citep{vishwakarma2025snorkelspatial}. Grid-Bench
evaluates complete-state prediction across eight operation families and
two-, three-, and four-dimensional grids. Its cell-level comparisons connect
exact recovery to the location and extent of differences in the returned
configuration.

\begin{table}[!htbp]
\centering
\small
\begingroup
\rowcolors{2}{TableStripe}{white}
\setlength{\tabcolsep}{3.5pt}
\begin{tabularx}{\textwidth}{>{\raggedright\arraybackslash}p{0.17\textwidth}*{4}{>{\raggedright\arraybackslash}X}}
\tablehead
\textbf{Work} & \textbf{Setting} & \textbf{Supplied information} & \textbf{Evaluated output} & \textbf{Evaluation focus} \\
\midrule
SPARTQA \citep{mirzaee2021spartqa} & Textual scenes & Description and question & Relation, object, or containment answer & Spatial relations and reasoning steps \\
CLEVR/\allowbreak NLVR/\allowbreak VSR \citep{johnson2017clevr,suhr2017nlvr,liu2023vsr} & Visual scenes & Image and language & Answer or statement truth & Compositional and spatial relations \\
SCONE \citep{long2016scone} & Executable domains & Initial world and instructions & Executable logical forms & Final-world denotation accuracy \\
ProPara/OpenPI \citep{dalvi2018propara,tandon2020openpi} & Natural processes & Procedural text & Entity states or change tuples & Changes through process steps \\
Entity tracking \citep{kim2023entity} & Containers and objects & Initial contents and events & Queried container contents & Update and retention \\
Interactive worlds \citep{cote2018textworld,shridhar2021alfworld,wang2022scienceworld} & Textual or embodied environments & Observations and goals & Agent-selected actions & Goal success and efficiency \\
SnorkelSpatial \citep{vishwakarma2025snorkelspatial} & Textual boards & Board, actions, and query & Final-property answer & Transformations and action count \\
Alien Grid \citep{chapra2026complexity} & $3\times3$ grids & Grid and event history & Intermediate and final grids & Sequential depth and failure types \\
Chess-World-Model \citep{walker2026chess} & Chess & Legal move prefix & 75 state labels per prefix & Exact and labelwise state accuracy \\
Grid-Bench & Multidimensional grids & Initial state, rules, and prescribed events & Complete final grid & Histories, spatial conditions, and cell errors \\
\bottomrule
\end{tabularx}
\endgroup
\caption{Representative evaluation settings related to Grid-Bench. The columns
describe what information a task supplies, what prediction or action it scores,
and the main focus of that evaluation. Rows group closely related datasets when
they share a role in the discussion; the table is not a numerical comparison.}
\label{tab:related}
\end{table}

\subsection{Sequential computation and structured reasoning}
\label{app:related-computation}
Formal and synthetic tasks study how neural models realize computations over
sequences. Computation-graph analyses cover multiplication, logic-grid tasks,
and dynamic programming \citep{dziri2023faith}. Automata experiments show that
the number of state transitions in a specification need not determine the
computation a Transformer learns \citep{liu2023automata}; evaluations across the
Chomsky hierarchy relate length generalization to task and architecture classes
\citep{deletang2023chomsky}. Permutation-composition studies combine behavioral
evaluation with probes and activation patching to examine learned algorithms
\citep{li2025state}. Scratchpads instead expose intermediate computations as
generated text for tasks including program execution \citep{nye2021scratchpads}.
In multi-hop question answering, \citet{wan2026fano} analyze single-pass
reasoning through an information-theoretic capacity model and use it to
motivate decomposition of interdependent evidence.

Program analysis provides a setting in which intermediate claims have formal
semantics. \citet{mitchell2025abstract} prompt LLMs to reason as abstract
interpreters, evaluating the soundness of generated invariants and recurring
reasoning errors. For natural-language queries, \citet{qin2025premise} instead
convert premises into logical representations and check them against retrieved
sources before generating a response.

Graph-based methods expose another range of structural dependencies. NQE
combines neural representations of n-ary facts with logical query operations
\citep{luo2023nqe}. GT Bench evaluates graph algorithms across graph structures
and input representations, and GTA uses representation selection and task
decomposition to improve performance \citep{xu2026gta}. CDCR-SFT trains models
to construct causal graphs and reason over variable-level dependencies
\citep{li2025causal}. TAG-AD evaluates
contextual and structural anomalies in text-attributed graphs
\citep{xu2025tagad}; LLM-GOOD combines LLM filtering with graph neural networks
for data-efficient out-of-distribution detection \citep{xu2025graph_ood}.
ReMindRAG uses graph traversal and replay of earlier traversals to retrieve
relevant knowledge \citep{hu2025remindrag}. These tasks differ in their outputs,
but all make the organization of information part of the reasoning problem.

Executable problems also support evaluation beyond a binary task outcome.
AlgoWorlds distinguishes feasible decisions from globally optimal ones using
exact solutions to combinatorial problems \citep{xu2026algoworlds}. QUITE
evaluates SQL rewrites for semantic equivalence and execution efficiency,
using database feedback during rewriting \citep{song2025quite}. TerminalTraj
constructs agent trajectories in Dockerized environments and pairs them with
executable validation code \citep{wu2026terminaltraj}.

A related line changes how reasoning is organized or learned. DyFlow constructs
and revises workflows from intermediate feedback \citep{wang2025dyflow};
CogWriter combines hierarchical planning with monitoring and revision for
constrained long-form generation \citep{wan2025cogwriter}. CoAct uses online
task allocation to parallelize LLM-agent execution \citep{peng2026coact}.
Context-CoT synthesizes and filters reasoning traces grounded in supplied
context for fine-tuning \citep{jin2026contextcot}, while AdaThinkV learns when
explicit reasoning improves video answers enough to justify its additional
response length \citep{tian2026adathinkv}.

Grid-Bench evaluates the complete spatial state produced by a supplied
transition sequence. History length counts events in this specification;
task-local reads and changes describe reference execution. These quantities
organize the conditions under which predictions are compared. The formal sequence
studies above address how models realize sequential mappings, complementing
the output-level evaluation of spatial states.

\subsection{Contextual robustness and evaluation}
\label{app:related-context}
Controlled changes to an input provide another way to characterize model
behavior. Adaptive Distraction uses tree search to construct task-irrelevant
context while preserving the original question and answer
\citep{wang2025distraction}. JevOut optimizes natural context additions to
redirect decisions under answer-preservation constraints \citep{xu2026jevout}.
Query-agnostic attacks on LLM retrieval show that document edits can also
change which evidence reaches a model \citep{li2026hidden}.
Cross-Lingual Pitfalls searches for bilingual questions that expose
performance discrepancies \citep{xu2025crosslingual}, and BICAUSE uses aligned
Chinese and English causal statements to examine sensitivity to language
and causal word order \citep{wang2025babel}. These studies vary the
presentation surrounding a reasoning problem; Grid-Bench varies the supplied
events and spatial configurations that determine its reference outcome.

Evaluation design also determines which aspects of a system become observable.
AutoBench-V generates images and visual questions to construct evaluations for
requested capabilities \citep{bao2024autobenchv}. AD-LLM separates anomaly
detection, data augmentation, and model selection into distinct tasks
\citep{yang2025adllm}. Within logical reasoning, FineLogic distinguishes final
accuracy from stepwise soundness and representation-level evidence
\citep{zhou2025finelogic}. Agent evaluations make similar distinctions:
FORTIS separately checks skill selection and execution within permitted
actions \citep{li2026fortis}, while Counterfactual Trace Auditing compares
paired trajectories to identify behavioral changes that pass rates can miss
\citep{zhou2026trace}.

Representational analyses and interventions address different questions from
output scoring. BiasLens measures associations among concepts in model
representations \citep{gao2025biaslens}; Treble Counterfactual VLMs uses causal
interventions to study modality shortcuts associated with hallucination
\citep{li2025treble}. Inside the Unfair Judge similarly studies scoring biases
through representations and interventions in LLM judges
\citep{xu2026unfairjudge}. Work on the checkability boundary places candidate
generation with the LLM and acceptance with an independent semantic checker
\citep{zhou2026checkability}. In Grid-Bench, reference execution supplies this
check: each returned cell is compared directly with its reference value.

\section{Benchmark Specification}
\label{app:tasks}
\subsection{Shared conventions and read accounting}
\label{app:task-conventions}
The state domain and finite alphabet are those defined in
Section~\ref{sec:state}. Each event supplies its operation and parameters.
Coordinates omitted from the input list have value zero, so the list specifies
the complete world.

Let $N_1(x)$ contain the in-bounds coordinates at Manhattan distance one from
$x$, and let $B_1(a,r)$ be the in-bounds Manhattan ball of radius $r$ centered
at $a$. Reference execution reads the current world $W$ and produces $W'$. Unless a rule
explicitly specifies sequential mutation, all reads use $W$. Cells not selected
for writing keep their values. The zero value is stored implicitly.
$\task{SET}(p,v)$ writes $v$ at $p$, including zero when a cell is cleared.

The canonical
read list $\mathcal R(W,e)$ counts cell incidences, so repeated accesses to
one coordinate can occur more than once. Its unique read support is
$R(W,e)=\{x:x\text{ occurs in }\mathcal R(W,e)\}$. A rule also specifies a
write support $S(W,e)$: the coordinates that receive assignments, whether or
not their values change. Finally,
\begin{align}
D(W,e)&=\{x:F_f(W,e)(x)\ne W(x)\},\\
\mathrm{scan}(W,e)&=|\mathcal R(W,e)|,\\
\mathrm{effect}(W,e)&=|D(W,e)|.
\label{eq:read-effect}
\end{align}
Thus $D(W,e)\subseteq S(W,e)$, but a large support need not produce a large
effect. A \task{SET} event has scan one and effect one when $W(p)\ne v$,
or effect zero otherwise. Reference reads use a fixed accounting convention.
Component boundary cells are counted once even if
several explored cells touch them; neighborhood aggregation instead counts each
incidence in a separate output sum.

The definitions below specify valid rule inputs and transition semantics.
Appendix~\ref{app:generation} gives the evaluated sampling conditions, and
Appendix~\ref{app:raw-rules} records the literal model-facing rule cards.

\subsection{Geometric selection and synchronous aggregation}
\label{app:geometric-rules}

\textbf{\task{PAINT}$(a,r,u,v)$.} For every $x\in B_1(a,r)$ with $W(x)=u$,
set $W'(x)=v$. Other values in the ball and all values outside it are unchanged.
Both $u=0$ and $v=0$ are valid, and $u=v$ is a valid no-op.
A radius of zero selects only the anchor. The canonical
read count is $|B_1(a,r)|$. The anchor is in bounds, $r$ is a nonnegative
integer, and both labels belong to the declared alphabet. The write support is
$\{x\in B_1(a,r):W(x)=u\}$, so selecting a ball does not imply recoloring
every cell in it. The source-value predicate is true when this set is nonempty.
The evaluated generator proposes only $u\ne v$, making predicate truth
equivalent to positive effect. Its no-op proposals select a source label
absent from the ball.

\textbf{\task{REFLECT}$(\mathrm{lo},\mathrm{hi},j)$.} Within the inclusive box
$\mathrm{lo}\leq x\leq\mathrm{hi}$, map coordinate $x$ to $x'$ with
$x'_j=\mathrm{lo}_j+\mathrm{hi}_j-x_j$ and other coordinates unchanged.
Set $W'(x')=W(x)$, including zero values, simultaneously for the entire box.
This is a bijection, so repeating the same reflection restores the input.
The read count is the box volume. A box of extent one on axis $j$ gives an
identity operation; generated reflections use at least two cells on that axis.
More generally, a reflection-symmetric box can be a no-op even when that extent
exceeds one. The lower and upper corners are in bounds and ordered coordinate
by coordinate. Both the read and write support are the entire inclusive box,
not only its occupied coordinates. Including zeros matters: reflecting a
nonzero value into an empty position also clears its old position.

\textbf{\task{STENCIL}$(a,r)$.} With alphabet $\{0,\ldots,M-1\}$, update each
$x\in B_1(a,r)$ by
\begin{equation}
W'(x)=\left(W(x)+\sum_{y\in N_1(x)}W(y)\right)\bmod M.
\label{eq:stencil}
\end{equation}
Neighbors outside the ball contribute to the sum but are not written. All
updates read the same old world, including when two updated cells are adjacent.
The workload target is update-support size; reference reads count
$\sum_{x\in B_1(a,r)}(1+|N_1(x)|)$, including repeated incidences.
The unique read support is the ball together with all of its in-bounds
neighbors. A cell outside the ball can consequently influence several output
cells while its own value is preserved. A zero-radius event still aggregates
the anchor's neighbors. The all-zero world is a fixed point. The generation
checks in Appendix~\ref{app:generator-algorithm} retain nonempty trajectories
and impose a minimum effect on sampled aggregation events.

\subsection{State-dependent regions and boundaries}
\label{app:connected-rules}
Here adjacency always means an in-bounds axis-neighbor edge, not diagonal
contact. For any set $C$, its external vertex boundary is
\begin{equation}
\partial C=\{y\in\domain\setminus C:\exists x\in C,\ y\in N_1(x)\}.
\end{equation}
Boundary coordinates are unique: a cell adjacent to several members of $C$
still contributes one element to $\partial C$.

\textbf{\task{MIX}$(p,r)$.} With alphabet $\{0,\ldots,M-1\}$ and $K=M-1\geq2$,
an empty anchor gives a no-op. Otherwise, find the set $S$ reachable from $p$
by at most $r$ axis-neighbor steps through cells with nonzero old values. Empty
cells cannot be entered or crossed. Simultaneously, for each $x\in S$, set
\begin{equation}
W'(x)=1+\left(W(x)+\sum_{y\in N_1(x)}W(y)\right)\bmod K.
\label{eq:mix}
\end{equation}
The sum includes neighbors outside $S$; empty neighbors contribute zero. The
nonzero support is conserved by the core event, but interleaved \task{SET}
operations may change occupancy. The target is $|S|$; reference reads count
$\sum_{x\in S}(1+|N_1(x)|)$, or one for an empty anchor. Unlike
\task{STENCIL}, this rule changes both support discovery and arithmetic.
The region is a graph-distance ball in the occupied subgraph, not all occupied
cells within a Manhattan ball. A coordinate can be geometrically close but
unreachable within $r$ steps because intervening empty cells block every short
path. At radius zero only the anchor is updated, although its neighbors are
still read. The selected generator proposes nonzero anchors and targets the
size of this reached region (Appendix~\ref{app:generation}).

\textbf{\task{RELABEL}$(p,v)$.} Find the maximal axis-connected component $C$
containing $p$ whose cells have value $W(p)$, and set every cell in $C$ to $v$.
The rule permits an empty-component anchor, deletion by $v=0$, and an unchanged
label. The read count is
$|C|+|\partial C|$, where $\partial C$ is the in-bounds external vertex boundary.
The component is identified entirely in the old world, so touching a component
of the destination label after recoloring does not expand this event's write
support. The next event can, however, encounter the combined component.
Appendix~\ref{app:generation} specifies the narrower proposal policy used
to sample these valid transitions.

\textbf{\task{LAYER}$(p,s)$.} An empty anchor is a no-op. Otherwise let $C$ be
the seed's maximal equal-valued component with value $u$. For $s=-1$, erase
every cell of $C$ with an in-bounds empty neighbor. For $s=+1$, fill every empty
cell of $\partial C$ with $u$. Reads are simultaneous. Out-of-world positions
are not empty neighbors: growing $[0,1,0]$ yields $[1,1,1]$, and subsequent
erosion leaves it unchanged. Growth and erosion are therefore not inverses.
Reference reads count $|C|+|\partial C|$, or one at an empty anchor.
For erosion, the write support is
$\{x\in C:N_1(x)\cap W^{-1}(0)\ne\varnothing\}$; for growth it is
$\partial C\cap W^{-1}(0)$. Erosion can split a component, whereas growth can
join same-label components through a newly filled cell. Neither operation
overwrites another nonzero label. Both signs have positive effect precisely
when the old external boundary contains an empty cell. The reference discovers
the component and its boundary before writing, so a newly filled cell does not
trigger another layer of growth within the event.

\subsection{Ordered motion and line aggregation}
\label{app:transport-rules}
\textbf{\task{MOVE}$(v,j,s,k)$.} Select the original entities of nonzero value
$v$. For direction $s=+1$, process them in decreasing coordinate on axis $j$;
for $s=-1$, use increasing order. Each moves at most $k$ one-cell steps,
stopping before a boundary or a currently occupied cell. Earlier moves are
visible to later ones. Each move clears its origin and writes $v$ at its
destination, without merging. Different axis-parallel lines are independent.
The write support also includes intermediate coordinates traversed and later
cleared during an entity's move, even when their final value equals their
initial value.
The read count is the number of processed entities plus all in-bounds
destination checks, including checks that encounter an obstacle.
The direction is $s\in\{-1,+1\}$ and $k$ is a nonnegative integer. At $k=0$
the scan is the selected entity count and the state is unchanged. An entity
already at the leading wall makes no in-bounds destination check; an entity
immediately behind an obstacle makes one. Original entities are processed once:
moving an entity into another coordinate does not add it again to the processing
list. Ties on the moving-axis coordinate occur only on distinct lines and do
not affect the result. The selected generator uses
$1\le k\le\min(4,L_j-1)$ rather than treating $k$ as the workload parameter.

\textbf{\task{COMPRESS}$(j,s)$.} On each line parallel to axis $j$, list the
nonzero values starting at the destination edge. Read this list in order,
replacing each adjacent equal pair by its sum and continuing after the pair.
Write the resulting list from that edge and set the remaining line to zero.
A newly merged value cannot merge again during the same event. For example,
$[2,2,4]$ toward coordinate zero becomes $[4,4,0]$, not $[8,0,0]$; a second
event produces the latter. Core events conserve each line's sum. Generated
values are zero or powers of two through $2^{12}$; an out-of-alphabet merge
is invalid and rejected by replay validation. Every core event reads the
entire world, making world size its workload parameter.
The order is stable before pairing: zeros are removed without sorting the
remaining values. The one-pass pairing rule also distinguishes
$[4,2,2,0]\mapsto[4,4,0,0]$ from an unrestricted repeated-merging process.
Opposite directions can choose different pairs when there are three equal
tiles. A core event cannot create an additional nonzero tile, but the
interleaved \task{SET} policy can inject small tiles, so the core conservation
properties are not claims about the whole event sequence.

\subsection{Checked illustrative examples}
\label{app:examples}
Table~\ref{tab:examples} illustrates each rule on a bounded line of five cells;
coordinates are $[0],\ldots,[4]$. These are explanatory examples, not additional
evaluated instances or human baselines. Expected outputs were specified directly
and checked against both the reference simulator and a separate dictionary-based
implementation. The explicit expected states make boundary and ordering
conventions directly checkable.

\begin{table}[!htbp]
\centering
\small
\begingroup
\rowcolors{2}{TableStripe}{white}
\begin{tabularx}{\textwidth}{lXXX}
\tablehead
\textbf{Rule} & \textbf{Initial values} & \textbf{Event parameters} & \textbf{Final values} \\
\midrule
\task{PAINT} & $[0,1,1,2,0]$ & $a=[2],r=1,u=1,v=3$ & $[0,3,3,2,0]$ \\
\task{REFLECT} & $[1,0,2,3,0]$ & $\mathrm{lo}=[1],\mathrm{hi}=[4],j=0$ & $[1,0,3,2,0]$ \\
\task{STENCIL} & $[1,2,0,3,0]$ & $a=[2],r=1,M=4$ & $[1,3,1,3,0]$ \\
\task{MIX} & $[1,2,0,3,1]$ & $p=[0],r=2,M=4$ & $[1,1,0,3,1]$ \\
\task{RELABEL} & $[1,1,0,1,2]$ & $p=[0],v=2$ & $[2,2,0,1,2]$ \\
\task{LAYER} & $[0,1,1,0,2]$ & $p=[1],s=+1$ & $[1,1,1,1,2]$ \\
\task{MOVE} & $[1,1,0,0,2]$ & $v=1,j=0,s=+1,k=2$ & $[0,0,1,1,2]$ \\
\task{COMPRESS} & $[2,2,4,4,0]$ & $j=0,s=-1$ & $[4,8,0,0,0]$ \\
\bottomrule
\end{tabularx}
\endgroup
\caption{Single-event illustrations with explicit expected states. All positions
not changed by the event retain their initial value. The \task{COMPRESS} alphabet
includes 8; other examples use an alphabet containing their displayed labels.}
\label{tab:examples}
\end{table}

Figure~\ref{fig:overview} shows all four reference states of two three-event
\task{MOVE} histories, each containing a \task{SET} assignment. In (A), coordinates are
$(\text{row},\text{column})$, with zero-based rows increasing downward and
columns rightward. The entity starts at $(3,0)$ and moves one cell right to
$(3,1)$. A \task{SET} then writes value 2 at $(1,1)$. The final \task{MOVE}
allows up to two cells upward but stops at $(2,1)$, immediately below the new
obstacle. In (B), coordinates are $(x,y,z)$. The first move follows $+x$ from
$(0,1,0)$ to $(1,1,0)$; \task{SET} writes 2 at $(1,1,2)$; the final move
follows $+z$ but stops at $(1,1,1)$. Existing obstacles retain their values
in both histories. The grids and exploded layers visualize explanatory
examples whose states are checked against both executors; model inputs use
the text serialization in Appendix~\ref{app:prompts}.

Figure~\ref{fig:technical-cases} adds four two-dimensional examples that expose
details hidden by a line illustration. Each panel's coordinates are
$[\text{row},\text{column}]$, with row increasing downward and column increasing
rightward from zero. Solid blue outlines mark write support in the old state;
dashed gray outlines mark other read coordinates. Orange outlines in the new
state identify actual value changes. A written cell need not change: the
\task{STENCIL} center keeps value 1, while in \task{MOVE} the original entity at
$[0,1]$ moves out and a following entity moves into that coordinate.
The motion panel also writes and clears the transit coordinates $[1,2]$ and
$[2,3]$; both are in the blue write support although they begin and end at zero.

\begin{figure}[!htbp]
\centering
\includegraphics[trim={8.75bp 4.35bp 11.6bp 10.25bp},clip,width=\linewidth]{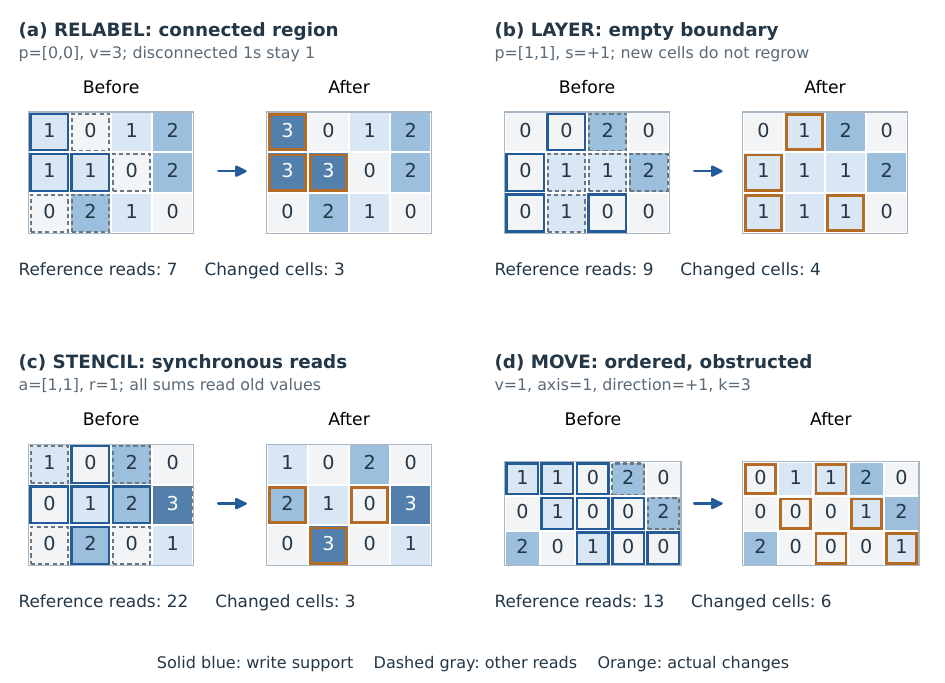}
\caption{Four hand-calculated two-dimensional rule examples, verified by the
production and dictionary executors. Cell color encodes the printed integer
value; solid blue old-state outlines mark write support, dashed gray outlines
mark other read coordinates, and orange new-state outlines mark actual changes.
\task{RELABEL} preserves disconnected cells with the same label.
\task{LAYER} grows only one boundary layer and does not overwrite label 2.
\task{STENCIL} reads beyond its five-cell update ball and uses only old values.
\task{MOVE} updates each row front to back, with label-2 obstacles preserved.
Read counts include repeated incidences, whereas outlines show unique
coordinates. These small examples are explanatory fixtures, not formal samples
or model predictions.}
\label{fig:technical-cases}
\end{figure}

In the \task{STENCIL} panel, the old-state sums on the five support coordinates,
in row order, are $4,2,5,8,3$. Modulo four these become $0,2,1,0,3$,
changing three values. Its 22 read incidences cover ten unique coordinates,
five outside the write support. In the motion panel, the first row's leading
entity advances one cell before encountering label 2; the following entity
then advances into its vacated coordinate. The middle row advances two cells,
and the bottom row advances to the world edge. These steps give four processed
entities plus nine in-bounds occupancy checks, not four times the maximum
distance.

\subsection{Geometry and transition properties}
\label{app:mathematics}
\label{app:geometry}
At nearly equal volume, the evaluated shapes have different adjacency and
boundary structure. Let $N=\prod_iL_i$ and let $E$ count undirected
axis-neighbor edges. Counting edges along each axis gives
\begin{align}
E&=\sum_{i=1}^{d}(L_i-1)\prod_{j\ne i}L_j,\\
\overline{\deg}&=\frac{2E}{N}
=2\sum_{i=1}^{d}\left(1-\frac1{L_i}\right).
\label{eq:geometry-degree}
\end{align}
The fraction of cells on the geometric world boundary is
\begin{equation}
b=1-\frac{\prod_i\max(L_i-2,0)}{N}.
\label{eq:boundary-fraction}
\end{equation}
Table~\ref{tab:technical-geometry} reports these quantities for the three
shapes. Higher dimension here increases mean degree and boundary exposure
while shortening individual axes. These are joint structural changes in
the dimension experiment.

\begin{table}[!htbp]
\centering
\small
\begingroup
\rowcolors{2}{TableStripe}{white}
\begin{tabularx}{\textwidth}{lXrrrrr}
\tablehead
$d$ & \textbf{Shape} & $N$ & $E$ & $\overline{\deg}$ & \textbf{Boundary} & $N+2E$ \\
\midrule
2 & $[16, 16]$ & 256 & 480 & 3.750 & 23.44\% & 1216 \\
3 & $[6, 6, 7]$ & 252 & 636 & 5.048 & 68.25\% & 1524 \\
4 & $[4, 4, 4, 4]$ & 256 & 768 & 6.000 & 93.75\% & 1792 \\
\bottomrule
\end{tabularx}
\endgroup
\caption{Exact structural quantities for the three evaluated shapes. Boundary is the percentage of cells on the geometric world boundary. The final column is full-world neighborhood-aggregation scan, not the realized workload of the selected STENCIL or MIX events.}
\label{tab:technical-geometry}
\end{table}

For a \task{STENCIL} or occupied-anchor \task{MIX} support $S$, reference reads
satisfy
\begin{equation}
\mathrm{scan}=\sum_{x\in S}(1+|N_1(x)|)\le(1+2d)|S|.
\label{eq:support-scan-bound}
\end{equation}
Full-world support gives $N+2E$. The count includes repeated neighborhood
incidences, so equal support sizes need not imply equal read counts.

\textbf{Conservation and ordering.}
\label{app:rule-properties}
The following properties hold for individual valid core events and provide
checks on execution.
Reflection permutes the inclusive box, preserving every label count; applying
the same reflection twice restores the state because its coordinate map is
an involution.
\task{MIX} preserves the occupied set: every updated value remains in
$\{1,\ldots,K\}$ and every updated coordinate was already occupied.
\task{MOVE} preserves label counts and ordering along each axis-parallel line.
If selected-label positions before and after the event are ordered as
$a_1<\cdots<a_m$ and $b_1<\cdots<b_m$, then
$0\le s(b_i-a_i)\le k$. Occupied cells prevent overtaking, while front-to-back
execution lets later entities enter vacated space.
\task{COMPRESS} preserves each line's sum: deleting zeros, replacing
$a,a$ by $2a$, and packing all preserve that sum. Its occupied-cell count
cannot increase. These event-specific identities do not extend unchanged
through arbitrary \task{SET} assignments.

\textbf{Synchronous aggregation.}
\label{app:aggregation-algebra}
Index the world lexicographically and write its values as
$w\in\mathbb Z_M^N$. For a fixed \task{STENCIL} event, let $A$ be the
grid adjacency matrix and $D_S$ the diagonal indicator of its update support.
Then
\begin{equation}
w'=(I+D_SA)w\pmod M.
\label{eq:stencil-matrix}
\end{equation}
Rows outside $S$ preserve their old values; rows inside add the old neighbors.
This is linear over the residue ring $\mathbb Z_M$, including the evaluated
$\mathbb Z_4$. Sequential in-place updates would define a different map.
\task{MIX} instead reduces modulo $K=M-1$ and adds one: on a fully updated
three-cell line $[1,1,1]$, the two rules yield $[2,3,2]$ and $[3,1,3]$,
respectively. Their arithmetic differs as well as their support selection.

\textbf{Event changes and final changes.}
\label{app:history-definitions}
The cumulative effect $\sum_{t=1}^T|D(W_{t-1},e_t)|$ counts a coordinate
again whenever it changes. The set $C_i$ in
Equation~\ref{eq:change-partition} instead compares only the initial and final
values. A cell can change repeatedly and still finish in $U_i$.
Reference workload and final-state diagnostics therefore measure distinct
properties of the same history.

\subsection{Generation conditions and acceptance}
\label{app:generation}
\label{app:generator-algorithm}
The formal evaluation uses generator 0.5.0 and the settings in
Table~\ref{tab:generation}. Initial states and event proposals are sampled
within each condition, then executed to obtain reference trajectories.
The main shapes are $[16,16]$, $[6,6,7]$, and $[4,4,4,4]$.
The workload sweep fixes $d=2$ and uses \task{PAINT} targets 25/100,
\task{REFLECT} 16/64/256, \task{LAYER} 15/60/150, and \task{MIX} 6/12/30.
Four levels coincide with main conditions and reuse those instances.

\begin{table}[!htbp]
\centering
\small
\begingroup
\rowcolors{2}{TableStripe}{white}
\begin{tabularx}{\textwidth}{lrlXrr}
\tablehead
\textbf{Rule} & \textbf{Target} & \textbf{Unit} & \textbf{Alphabet / initializer} & \textbf{Density band} & \textbf{SET rate} \\
\midrule
\task{PAINT} & 100 & reads & $0{:}3$ / uniform & .45--.65 & .2 \\
\task{REFLECT} & 64 & reads & $0{:}3$ / uniform & .45--.65 & .2 \\
\task{STENCIL} & 10 & support & $0{:}3$ / uniform & .55--.85 & .2 \\
\task{MIX} & 12 & support & $0{:}3$ / uniform & .55--.70 & .2 \\
\task{RELABEL} & 100 & reads & $0{:}6$ / regions & .50--.65 & .2 \\
\task{LAYER} & 60 & reads & $0{:}6$ / regions & .45--.60 & .2 \\
\task{MOVE} & 30 & reads & $0{:}4$ / uniform & .35--.50 & .3 \\
\task{COMPRESS} & 256 & cells & $\{0,2^0,\ldots,2^{12}\}$ / small tiles & .15--.60 & .5 \\
\bottomrule
\end{tabularx}
\endgroup
\caption{Main generation conditions. Targets guide candidate selection;
realized operation sizes vary with the state and geometry. A cells target
selects the shape tier, whose volume is 252 or 256. Density bands guide
initialization and event selection.}
\label{tab:generation}
\end{table}

\textbf{Initial states.} A density $\rho$ is drawn uniformly from the family's
band $[l,h]$, giving $\max(1,\min(N-1,\operatorname{round}(N\rho)))$ occupied
cells. Uniform initialization samples coordinates without replacement and
assigns independent nonzero labels. Small-tile initialization uses the two
smallest positive tile values with probabilities 0.9 and 0.1.
Region initialization grows connected sets through empty axis neighbors,
choosing their typical size to approximate the component-plus-boundary target.
Regions receive a label absent from their external neighbors when possible.
The latter initializer allows workload to vary both initial component structure
and later operations.

\textbf{Event schedule.} At rate $s$, position $t$ is a \task{SET} event when
$\lfloor ts\rfloor>\lfloor(t-1)s\rfloor$. The first event is therefore a core
operation. Assignments clear cells above the density band, fill below it, and
otherwise change a value. The region families prefer nonzero labels distinct
from neighboring labels; \task{COMPRESS} spawns tiles of value 1 or 2, with
probabilities 0.9 and 0.1, and clears a cell when occupancy is too high.
At $T=64$, most families contain 52 core events and 12 assignments;
\task{MOVE} contains 45 and 19, and \task{COMPRESS} 32 and 32.
Assignments enter the supplied history because they can alter the state
used by subsequent operations. Reference reads and changed-cell counts
separately describe operation scale.

\textbf{Candidate selection.} Valid candidates are executed before selection.
For realized workload $q$, target $w$, tolerance $\delta$, and post-event
density $\rho'$, the selection score is
\begin{align}
p_w(q)&=\left[\left|\log\frac{\max(q,1)}{\max(w,1)}\right|
-\log(1+\delta)\right]_+,\\
p_\rho(\rho')&=\frac{2}{h-l}
\left[\left|\rho'-\frac{l+h}{2}\right|-\frac{h-l}{2}\right]_+,\\
\operatorname{score}(e)&=p_w(q_e)+p_\rho(\rho'_e).
\label{eq:candidate-score}
\end{align}
Here $q$ is the family's read count, support size, or world volume.
Tolerance is 0.15 for \task{PAINT}/\task{MIX}, 0.30 for \task{STENCIL},
0.25 for \task{REFLECT}/\task{RELABEL}/\task{LAYER}/\task{MOVE}, and zero
for \task{COMPRESS}. The minimum-score candidate is selected with seeded
tie-breaking. Before scoring, recent-state repetition is filtered when a fresh
alternative exists, and the requested effective/no-op category is preferred.
Applicable families target a 0.1 no-op proposal rate in shuffled blocks of
eight core events; \task{STENCIL}, \task{MIX}, and \task{RELABEL} request
effective events throughout. Unavailable categories are recorded as misses.

\textbf{Trajectory acceptance.} Generated states must remain nonempty and
within density bounds $[\max(0.02,l-0.4),\min(0.98,h+0.4)]$.
The mean-workload ratio must lie in $[0.5,2]$, the effective-core fraction
must be at least 0.5, and the missed-category fraction at most 0.5.
\task{STENCIL}/\task{MIX} core events must change at least 10\% of their support.
Up to eight seeded attempts are made per trajectory index. The stored
trajectories extend to $T=128$; acceptance is applied before selecting the
evaluated prefixes $T=0,1,8,16,32,64$. Dataset seed 20260907 and a
condition/index/attempt-specific digest determine the random streams.
The two collection cohorts select disjoint indices 0--3 and 4--7 from
the same conditions. Exact proposer and seed specifications accompany the
implementation.

\textbf{Realized operation scale.} Targets describe sampling objectives.
For example, at \task{MIX} targets 6/12/30, mean reads per core event in the
selected $d=2,T=64$ histories are 28.13/58.67/145.47, while mean changed-cell
counts are 3.99/8.00/20.05. Reads count repeated neighborhood incidences;
changes count coordinates whose values actually change. The supplementary
data provide these quantities for every main and additional workload condition.

\textbf{Connectivity profile.}
\label{app:relabel-distribution}
\task{RELABEL} proposals prefer destination labels absent from the component's
external boundary, using deletion or a merging label when necessary.
Deletion is eligible when the component occupies at most one quarter of the
nonzero cells and is tried first with probability 0.15.
The 24 main trajectories contain 54 contact events among 1,248 core
recolorings (4.3\%); 17 trajectories contain at least one.
This describes the low-merger distribution used for the task profile.
Assignments can also alter connectivity. Per-trajectory contact counts are
included in the supplementary data.

\section{Evaluation Protocol}
\label{app:protocol}
\subsection{Shared evaluation and model settings}
\label{app:collection}
The six models receive the same 1,488 instances from 248 reference
trajectories, producing 8,928 retained outcomes. The main experiment contains
192 trajectories and 1,152 instances. Seven additional workload conditions
contribute 56 trajectories and 336 instances; combining them with four reused
main conditions gives the 528-instance workload view. Each generation condition
has eight trajectories, and each trajectory supplies six separately queried
prefixes. At one main length, task and dimension summaries contain 24 and 64
instances, respectively.

Table~\ref{tab:endpoints} records the full model identities and request
parameters. GPT-5.6 Sol, Claude 5 Sonnet, and GPT-5.4 Mini use the NAIRR
gateway; GLM-5, Kimi K2.5, and MiniMax M2.5 use native Bedrock streaming
\citep{aws2026glm,aws2026kimi,aws2026minimax}.
Collection took place on 16--24 September 2026 UTC.
All requests use prompt 0.4, one completion, a 128,000-token combined
reasoning-and-answer ceiling, no tools, and a 7,200-second deadline.
Temperature, \texttt{top\_p}, and seed are omitted.
Five models receive the high-effort setting; MiniMax M2.5 uses native
always-on thinking without an effort field.

\begin{table}[!htbp]
\centering
\small
\begingroup
\setlength{\tabcolsep}{4pt}
\renewcommand{\arraystretch}{1.08}
\begin{tabularx}{\textwidth}{l>{\hsize=.8\hsize\linewidth=\hsize\raggedright\arraybackslash}X>{\hsize=.9\hsize\linewidth=\hsize\raggedright\arraybackslash}X>{\hsize=1.3\hsize\linewidth=\hsize\raggedright\arraybackslash}X}
\toprule
\tablehead
Model & Requested identity & Recorded route & Recorded request parameters \\
\midrule
\rowcolor{TableStripe}
GPT-5.6 Sol & \nolinkurl{gpt-5.6-sol} & \nolinkurl{azure/gpt-5.6-sol} & \nolinkurl{max_completion_tokens} = \texttt{128000}\newline \nolinkurl{reasoning_effort} = \texttt{"high"} \\
\rowcolor{white}
Claude 5 Sonnet & \nolinkurl{claude-sonnet} & \nolinkurl{vertex_ai/claude-sonnet-5} & \nolinkurl{max_completion_tokens} = \texttt{128000}\newline \nolinkurl{reasoning_effort} = \texttt{"high"} \\
\rowcolor{TableStripe}
GPT-5.4 Mini & \nolinkurl{gpt-5.4-mini} & \nolinkurl{azure/gpt-5.4-mini} & \nolinkurl{max_completion_tokens} = \texttt{128000}\newline \nolinkurl{reasoning_effort} = \texttt{"high"} \\
\rowcolor{white}
GLM-5 & \nolinkurl{zai.glm-5} & \nolinkurl{zai.glm-5} & \nolinkurl{max_tokens} = \texttt{128000}\newline \nolinkurl{reasoning_effort} = \texttt{"high"} \\
\rowcolor{TableStripe}
Kimi K2.5 & \nolinkurl{moonshotai.kimi-k2.5} & \nolinkurl{moonshotai.kimi-k2.5} & \nolinkurl{max_tokens} = \texttt{128000}\newline \nolinkurl{reasoning_effort} = \texttt{"high"} \\
\rowcolor{white}
MiniMax M2.5 & \nolinkurl{minimax.minimax-m2.5} & \nolinkurl{minimax.minimax-m2.5} & \nolinkurl{max_tokens} = \texttt{128000}\newline Effort absent; native thinking. \\
\bottomrule
\end{tabularx}
\endgroup
\caption{Six evaluated models, each with 1,488/1,488 scored instances, prompt 0.4, a 128,000-token generation allowance, and no tools. Parameter keys and values are copied from the retained request ledger. For Bedrock, ledger \texttt{max\_tokens} maps to wire field \texttt{inferenceConfig.maxTokens}; gateway requests retain \texttt{max\_completion\_tokens}. MiniMax M2.5 sends no effort field and uses native always-on thinking. Identifiers and routes follow the frozen collection metadata.}
\label{tab:endpoints}
\end{table}

Each instance is submitted as one user message, with no task-specific system
message or preceding conversation. Scoring uses the delivered answer text.
Technical failures before a scorable outcome can be retried; delivered
outcomes are retained regardless of correctness, format, or termination.
There is one retained outcome per model--instance pair. The request ledger
preserves exact parameters and raw termination fields.
\subsection{Model-facing prompts}
\label{app:prompts}
\label{app:prompt-template}
Figure~\ref{fig:technical-prompt-template} gives the shared prompt.
The space declaration specifies the shape, alphabet, and zero-based
non-wrapping coordinates. Initial states and events are compact JSON;
omitted cells are zero. Each prompt contains the complete event sequence
$e_1,\ldots,e_T$, including any assignments, and requests the complete state
after all $T$ events.

\begin{figure}[!htbp]
\centering
\begin{minipage}{\textwidth}
\small
\begin{verbatim}
This is a deterministic simulation of an abstract integer-valued grid.
Keep deliberation concise and reserve enough of the response budget to return the 
complete final-state JSON.

TASK
Apply all events in order and return the final world.

SPACE
shape = <axis lengths>
Coordinates and axis indices are zero-based. The world is bounded and does not wrap.
Unlisted cells have value 0. Allowed values: <declared alphabet>.
W(x) denotes the value of cell x immediately before the current event.

COMMON EVENT
SET(p,v): set cell p to v.

FAMILY RULE
<family-specific rule card>

EXAMPLE
<family-specific one-dimensional worked example>

INITIAL
<complete initial state in sparse JSON>

EVENTS
<ordered event array in JSON>

OUTPUT
Return only {"cells":[[p,v],...]}, where each entry is [coordinate,value]. Include 
every nonzero cell exactly once in lexicographic coordinate order. Output the JSON 
object only, with no code fence or other text.
\end{verbatim}
\end{minipage}
\caption{Shared prompt 0.4 template, derived from the actual renderer. Angle-bracket fields replace instance-specific content. Line wrapping is typographical; the evaluated representation remains a coordinate list, not a grid drawing.}
\label{fig:technical-prompt-template}
\end{figure}

\textbf{Rule cards.}
\label{app:raw-rules}
The following text is taken directly from the registered rule renderers
using the selected alphabets. The mathematical explanations in
Appendix~\ref{app:tasks} supplement these literal instructions for the reader.
\begingroup
\small
\textbf{PAINT.} PAINT(a,r,u,v): simultaneously replace u by v at every coordinate x with sum\_i |x\_i-a\_i| <= r. All other cells remain unchanged.

\textbf{REFLECT.} REFLECT(lo,hi,axis): for every cell x in the inclusive box lo <= x <= hi, the new value at the mirrored position x' (x'\_axis = lo\_axis + hi\_axis - x\_axis, other coordinates unchanged) is the old value at x, including 0. All writes happen simultaneously. Cells outside the box remain unchanged.

\textbf{STENCIL.} STENCIL(a,r): simultaneously set every cell x with sum\_i |x\_i-a\_i| <= r to (W(x) + sum W(y)) mod 4, where y ranges over the in-bounds cells at Manhattan distance 1 from x, whether or not they lie within distance r of a. Every update reads the same old world; cells farther than r from a remain unchanged.

\textbf{MIX.} MIX(p,r): if W(p) = 0, nothing changes. Otherwise find every cell that can be reached from p by a path of at most r Manhattan-distance-1 steps whose cells all have nonzero old values (p itself counts as reached with 0 steps; a cell with old value 0 can never be entered or passed through). Then simultaneously set every reached cell x to 1 + ((W(x) + sum W(y)) mod 3), where y ranges over the in-bounds cells at Manhattan distance 1 from x, whether or not they were reached. Every update reads the old world; cells that were not reached remain unchanged.

\textbf{RELABEL.} RELABEL(p,v): let u = W(p) (u may be 0). Find the maximal set of cells connected to p by Manhattan-distance-1 steps whose old values are all u, including p. Set every cell of that set to v.

\textbf{LAYER.} LAYER(p,s): if W(p) = 0, nothing changes. Otherwise let u = W(p) and find the maximal Manhattan-connected set of cells with old value u, including p. If s = -1, set to 0 every cell of that set that has an in-bounds neighbor with old value 0. If s = +1, set to u every in-bounds cell with old value 0 that is a neighbor of a cell of that set. Positions outside the world are not neighbors; nothing else changes.

\textbf{MOVE.} MOVE(v,axis,direction,k): take every cell with value v, ordered by decreasing coordinate on axis if direction=1 and increasing if direction=-1, so that the cells nearest the leading edge are handled first. In that order, move each one along axis in the sign of direction, one cell at a time and at most k cells, stopping as soon as the next cell is outside the world or nonzero (including cells already moved into). Each move clears the old cell and writes v to the new cell; values never merge, and a block of adjacent v cells therefore moves as a unit.

\textbf{COMPRESS.} COMPRESS(axis,direction): on every line parallel to axis, independently: list the nonzero values in order starting from the destination edge (coordinate 0 if direction=-1, the highest coordinate if direction=1). Walk that list from its start; whenever two consecutive values are equal, replace them by one value equal to their sum and continue after the pair, so a merged value cannot merge again in this event. Write the resulting list back starting at the destination edge and set every remaining cell of the line to 0.

\endgroup

\textbf{A complete example.}
\label{app:rendered-prompt}
Figure~\ref{fig:technical-worked-prompt} shows the full rendered prompt for
the \task{RELABEL} example in Figure~\ref{fig:technical-cases}.
It includes the renderer's one-dimensional rule example, followed by the
two-dimensional initial state and event.
The renderer computes the rule example under the instance's alphabet,
omitting it only when its values cannot be represented.
The evaluated instance's final state is never part of the prompt.

For this illustrative $3\times4$ world, recoloring the component at $[0,0]$
leaves the disconnected value-1 cells at $[0,2]$ and $[2,2]$ unchanged.
Figure~\ref{fig:technical-cases} shows the checked reference state;
the data package includes the literal prompt and answer.

\begin{figure}[!htbp]
\centering
\begin{minipage}{\textwidth}
\small
\begin{verbatim}
This is a deterministic simulation of an abstract integer-valued grid.
Keep deliberation concise and reserve enough of the response budget to return the 
complete final-state JSON.

TASK
Apply all events in order and return the final world.

SPACE
shape = [3,4]
Coordinates and axis indices are zero-based. The world is bounded and does not wrap.
Unlisted cells have value 0. Allowed values: [0,1,2,3].
W(x) denotes the value of cell x immediately before the current event.

COMMON EVENT
SET(p,v): set cell p to v.

FAMILY RULE
RELABEL(p,v): let u = W(p) (u may be 0). Find the maximal set of cells connected to p 
by Manhattan-distance-1 steps whose old values are all u, including p. Set every cell 
of that set to v.

EXAMPLE
On shape [7], listing the values of cells [0] to [6] in order. Before: 1 1 0 1 2 2 3. 
Event: {"op":"RELABEL","p":[4],"v":3}. After: 1 1 0 1 3 3 3. The cells with old value 2
 connected to [4] are [4] and [5]; both become 3. The 1 at [3] and the 3 at [6] are not
 part of that set, although [6] ends up adjacent to the new 3s.

INITIAL
{"cells":[[[0,0],1],[[0,2],1],[[0,3],2],[[1,0],1],[[1,1],1],[[1,3],2],[[2,1],2],[[2,2],
1]]}

EVENTS
[{"op":"RELABEL","p":[0,0],"v":3}]

OUTPUT
Return only {"cells":[[p,v],...]}, where each entry is [coordinate,value]. Include 
every nonzero cell exactly once in lexicographic coordinate order. Output the JSON 
object only, with no code fence or other text.
\end{verbatim}
\end{minipage}
\caption{Complete renderer-produced prompt for the checked RELABEL illustration in Figure~\ref{fig:technical-cases}. The 3-by-4 world is a synthetic explanatory fixture, not an evaluated formal instance. The included one-dimensional example is the renderer's standard rule example, not the answer to the supplied 2D world. Only printed line wrapping differs from the saved literal prompt.}
\label{fig:technical-worked-prompt}
\end{figure}

\subsection{Answer extraction and diagnostic populations}
\label{app:scoring}
\label{app:parser-examples}
The answer has one field, \texttt{cells}, listing every nonzero
coordinate--value pair. Omitted coordinates are zero. Duplicate or
out-of-bounds coordinates, explicit zeros, undeclared values, and unknown
fields invalidate a state. JSON parsing rejects duplicate keys and non-finite
numbers; Boolean values are not integer coordinates or labels.

The fixed extractor first tries the stripped response, optionally removing
one enclosing code fence. Otherwise, it selects the last top-level JSON-object
candidate, respecting braces inside strings. A malformed or unclosed trailing
candidate fails extraction. Selection uses syntax and position, never the
reference answer, and does not repair state values.
Semantic EM scores the extracted state. Strict EM applies the same scorer
directly to the raw response. For example, reordering a correct coordinate
list preserves both scores, while enclosing it in prose can preserve semantic
EM but fail strict EM. A valid state with one wrong cell has EM zero.

Section~\ref{sec:measurement} defines full-world EM, valid-state cell accuracy,
and the final changed/unchanged partition. These measures use three populations:
all requests for EM, valid states for cell accuracy, and valid incorrect
states for error extent and error categories.
For the valid subset $V$, define
$V_C=\{i\in V:|C_i|>0\}$ and $V_U=\{i\in V:|U_i|>0\}$.
The changed-cell accuracy is
\begin{equation}
\operatorname{CA}_C
=\frac{1}{|V_C|}\sum_{i\in V_C}
\left(1-\frac{|E_i\cap C_i|}{|C_i|}\right),
\label{eq:changed-ca}
\end{equation}
with the analogous definition on $U_i$. This averages per-instance fractions.
An empty contributing population has undefined accuracy; in particular,
$C_i$ is empty for the reconstruction control $T=0$.

Wrong-cell quantiles use linear interpolation at order position $(n-1)q$
for $n$ valid incorrect answers and quantile fraction $q$.
Empty subsets are shown as dashes. The $\leq5$-cell count describes small
errors without changing the exact-match criterion.
The complete error-partition counts and contributing populations accompany
the supplementary data.

\textbf{Answer accounting.}
\label{app:format-results}
\label{app:resource-results}
Correct, valid incorrect, nonempty invalid, and empty answers form a disjoint
partition of the full evaluation. A valid all-zero state,
\texttt{\{"cells":[]\}}, is a prediction rather than an empty response.
Termination is recorded separately; any delivered valid state is scored on its
content. Across the six models' 5,760 main responses at $T>0$, 1,457 are exact,
3,554 are valid but incorrect, and 749 are invalid or empty.
Table~\ref{tab:revision-history-summary} reports answer counts by
model and length, combining nonempty invalid and empty responses in its third
category. The data package retains their separate counts, semantic and strict EM,
and provider-reported usage for all six models. Request settings remain
specified here; the detailed records are indexed in Appendix~\ref{app:data-records}.

\subsection{Uncertainty and reference verification}
\label{app:statistics}
Intervals use 2,000 stratified bootstrap resamples of whole trajectories,
with pseudorandom seed zero. A stratum fixes family, dimension, and workload.
Resampling retains each stratum's eight trajectories and keeps the selected
prefix vector of a sampled trajectory intact. Thus the procedure preserves
dependence across lengths and compares models on matched instances.

Write $y_{srt}$ for correctness on trajectory $r$ in stratum $s$ at length $t$.
For selected lengths $\mathcal T$, the estimate is
\begin{equation}
\widehat\mu=
\frac{\sum_s\sum_{r=1}^{n_s}\sum_{t\in\mathcal T}y_{srt}}
{|\mathcal T|\sum_s n_s}.
\label{eq:cluster-mean}
\end{equation}
In each replicate, draw $n_s$ trajectory indices with replacement within
stratum $s$ and replace each trajectory by its sampled prefix vector.
The reported endpoints are the interpolated 2.5th and 97.5th percentiles.
For model differences, replace $y_{srt}$ by $y^A_{srt}-y^B_{srt}$ before
resampling. These are marginal intervals conditional on the evaluated
strata; degenerate empirical intervals are omitted.
Task, dimension, and workload breakdowns report descriptive counts.
Reused main/workload instances enter each selected view once.

Offline verification re-scores all 8,928 outcomes and checks the 1,488
reference states with separately implemented dictionary-based rules.
Both checks yield zero discrepancies, covering 30,008 replayed prefix events.
The rule illustrations and model-facing examples are also checked against
literal expected states. The implementation and technical-asset tests provide
the executable verification procedures.

\subsection{Supplementary data and reproduction}
\label{app:data-records}
\label{app:complete}
The public Hugging Face dataset provides benchmark instances, reference states,
and recorded measurements supporting the analyses in
Appendices~\ref{app:results} and~\ref{app:diagnostics}.
Its README defines units and populations, and its manifests record source
and member checksums. The \nolinkurl{evidence/} directory contains frozen
analysis and diagnostic summaries in JSON and CSV tables for convenient lookup.

Within \nolinkurl{evidence/}, the \nolinkurl{tables/} directory contains overall,
history-length, task, dimension, workload, and resource summaries.
The \nolinkurl{diagnostics/} directory contains joint task--dimension--length
counts, wrong-cell distributions, update/preservation partitions, matched
baselines, and paired-instance comparisons.
The \nolinkurl{reference/} directory contains realized workloads, omission
and suffix diagnostics, and per-trajectory connectivity counts.
Original and additional cohorts are identified explicitly.
The companion GitHub code replays reference states and rebuilds six descriptive
tables and the history-length plot from the released measurements.
Collected model answers are not distributed.

\section{Supplementary Performance Analyses}
\label{app:results}
These analyses resolve the aggregate results in Section~\ref{sec:experiments}
by task, dimension, workload, and collection cohort, then examine reversals
within individual histories. Appendix~\ref{app:data-records} indexes the
complete joint counts and model-pair records.

\FloatBarrier
\subsection{Recovery across histories and task families}
\label{app:results-context}
Table~\ref{tab:revision-history-summary} separates exact states, incorrect
grids, and invalid or empty responses by model and history length. The exact
counts underlie Figure~\ref{fig:length}. The reconstruction control and
the one-event condition play different roles: the former checks recovery
of the supplied state, while the latter already requires a spatial
transformation. All six models approach perfect reconstruction, but their
one-event scores range from 69 to 183 out of 192.

\begin{table}[!htbp]
\centering
\small
\begingroup
\setlength{\tabcolsep}{3pt}
\renewcommand{\arraystretch}{1.08}
\begin{tabularx}{\textwidth}{l*{6}{>{\centering\arraybackslash}X}}
\toprule
\tablehead
Model & $T=0$ & $T=1$ & $T=8$ & $T=16$ & $T=32$ & $T=64$ \\
\midrule
\rowcolor{TableStripe}
GPT-5.6 Sol & 192/0/0 & 183/9/0 & 145/47/0 & 87/105/0 & 57/135/0 & 25/166/1 \\
\rowcolor{white}
Claude 5 Sonnet & 192/0/0 & 137/54/1 & 91/98/3 & 61/111/20 & 35/102/55 & 14/44/134 \\
\rowcolor{TableStripe}
GPT-5.4 Mini & 191/1/0 & 69/120/3 & 18/157/17 & 5/166/21 & 2/153/37 & 1/150/41 \\
\rowcolor{white}
GLM-5 & 191/0/1 & 96/95/1 & 17/165/10 & 8/169/15 & 3/167/22 & 0/148/44 \\
\rowcolor{TableStripe}
Kimi K2.5 & 192/0/0 & 148/43/1 & 33/138/21 & 18/153/21 & 6/151/35 & 0/138/54 \\
\rowcolor{white}
MiniMax M2.5 & 188/2/2 & 150/40/2 & 31/117/44 & 11/133/48 & 5/144/43 & 1/136/55 \\
\bottomrule
\end{tabularx}
\endgroup
\caption{Answer composition at each main history length. Each cell lists exact / valid incorrect / invalid or empty counts, which sum to 192. The first two categories form the valid-state population; the second is used for error-extent and location diagnostics. All six lengths use separately queried prefixes of the same trajectories.}
\label{tab:revision-history-summary}
\end{table}

\begin{figure}[!htbp]
\centering
\includegraphics[trim={3.5bp 3.86bp 0.6bp 4.5bp},clip,width=\linewidth]{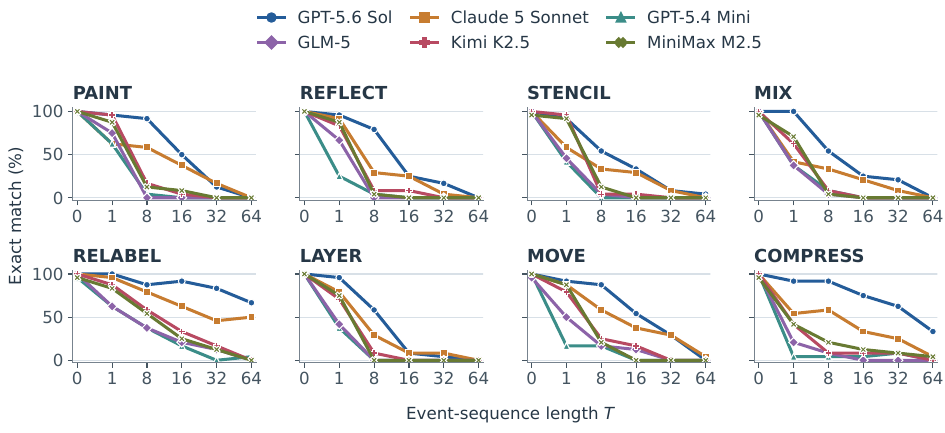}
\caption{Full task-specific history profiles. Each point is exact answers
out of 24 main instances: eight trajectories in each of three dimensions.
Colors and markers identify the six models. Lines connect the tested
categorical lengths; the curves are descriptive. All outcomes remain in
the denominator.}
\label{fig:task-curves}
\end{figure}

Figure~\ref{fig:task-curves} resolves the aggregate decline by task family.
The profiles differ in both one-event accuracy and the lengths over which
success persists. For GPT-5.6 Sol, \task{PAINT} falls from 22/24 at $T=8$
to zero at $T=64$, while \task{RELABEL} and \task{COMPRESS} retain 16/24
and 8/24. Claude 5 Sonnet's longest-history successes likewise concentrate
in \task{RELABEL}, with one each in \task{MOVE} and \task{COMPRESS}.
The \task{RELABEL} results use the low-merger generation
conditions specified in Appendix~\ref{app:relabel-distribution}.

\FloatBarrier

Within a task family, success need not decrease at every sampled length.
GPT-5.6 Sol's \task{RELABEL} count rises from 21 to 22
between $T=8$ and $T=16$, and Claude 5 Sonnet's rises from 11 to 12
between $T=32$ and $T=64$. Appendix~\ref{app:prefix-patterns} examines
such reversals on individual trajectories.

\FloatBarrier
\subsection{Spatial dimension across history lengths}
\label{app:structural-context}
\FloatBarrier
\begin{figure}[!htbp]
\centering
\includegraphics[trim={6.5bp 4.95bp 4.1bp 9bp},clip,width=\linewidth]{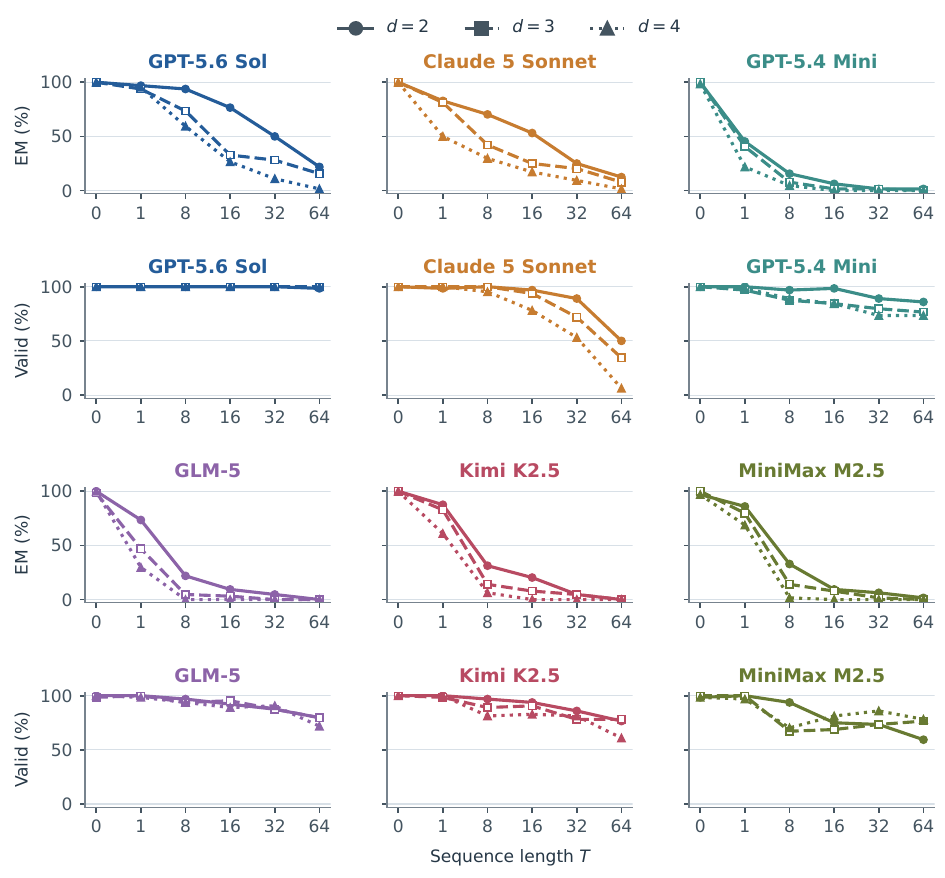}
\caption{Dimension profiles for all six models. Each model has two vertically
adjacent panels: full-world EM and valid-state rate. Circles/solid,
squares/dashed, and triangles/dotted denote $d=2,3,4$.
Each point includes 64 main instances from eight tasks and eight trajectories
per task. Valid states include exact and incorrect answers. Lines connect
categorical lengths and carry no uncertainty bands.}
\label{fig:structure}
\label{fig:dimensions-full}
\end{figure}

Figure~\ref{fig:structure} shows that dimension differences emerge at
different lengths across models. GPT-5.6 Sol's nearly level one-event
profile separates by eight events, when it solves 60/64, 47/64, and 38/64
instances across $d=2,3,4$. GLM-5 and Kimi K2.5 already show substantial
differences at one event, as reported in Table~\ref{tab:dimension-summary}.
Pooled over $T=8,16,32,64$, correct counts are nonincreasing from $d=2$ to
$d=4$ for every model.

At later lengths several exact-match curves approach zero, while their
companion valid-state curves remain above zero. Thus, many responses can
still be inspected cell by cell after complete recovery has become rare.
The dimension comparison combines changes in adjacency, boundary exposure,
and coordinate arity at approximately fixed volume; their geometric
quantities are given in Appendix~\ref{app:geometry}.

\FloatBarrier
\subsection{Within-family workload profiles}
\label{app:workload-profiles}
The workload curves retain all six lengths for the eleven settings at $d=2$.
Each point contains eight trajectories; each pooled main-text count contains
four prefixes of those trajectories.

\FloatBarrier
\begin{figure}[H]
\captionsetup{width=0.85\linewidth}
\centering
\includegraphics[trim={4.25bp 4.55bp 3.6bp 7bp},clip,width=0.85\linewidth]{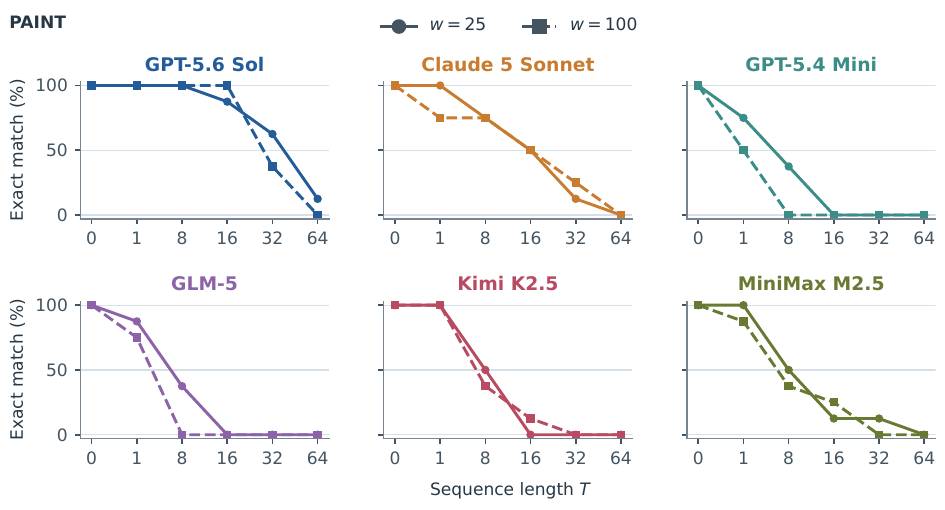}
\caption{\task{PAINT} workload profiles at $d=2$. Six model panels compare
the two ball-size targets. Each point is EM over eight trajectories at the
specified length, including all outcomes. Markers and line styles identify
targets; lengths are categorical. The main-condition level reuses main instances.}
\label{fig:workloads-full}
\end{figure}

The mixed pooled \task{PAINT} results include changes in ordering across
history lengths. GPT-5.6 Sol solves 7/8 and 8/8 instances at $T=16$ for
ball-size targets 25 and 100, respectively. At $T=32$, it solves 5/8 and
3/8, reversing that ordering. A single pooled comparison thus combines
different length-specific contrasts.

\FloatBarrier
\begin{figure}[H]
\captionsetup{width=0.85\linewidth}
\centering
\includegraphics[trim={4.25bp 4.55bp 3.6bp 6.75bp},clip,width=0.85\linewidth]{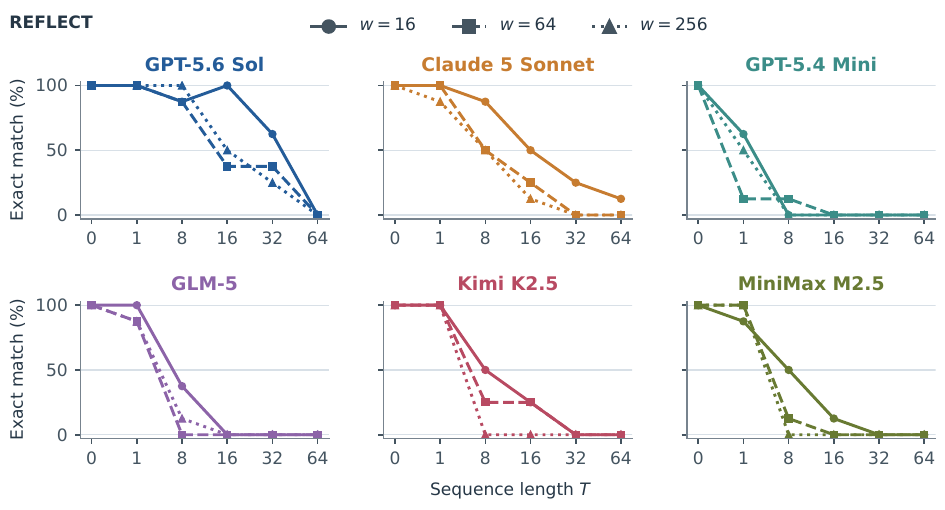}
\caption{\task{REFLECT} workload profiles at $d=2$. Each model panel compares
box-volume targets 16, 64, and 256 using the legend's markers and line styles.
Each point is EM over eight trajectories at a categorical tested length,
with all outcomes retained. Curves are descriptive.}
\label{fig:workloads-full-2}
\end{figure}

For \task{REFLECT}, GPT-5.6 Sol solves all eight one-event instances at
each box-volume target. At $T=16$, its counts are 8/8, 3/8, and 4/8 at
targets 16, 64, and 256. The smallest box retains the highest recovery at this
length, but the three-target profile is not monotonic.

\FloatBarrier
\begin{figure}[H]
\captionsetup{width=0.85\linewidth}
\centering
\includegraphics[trim={4.25bp 4.55bp 3.6bp 7bp},clip,width=0.85\linewidth]{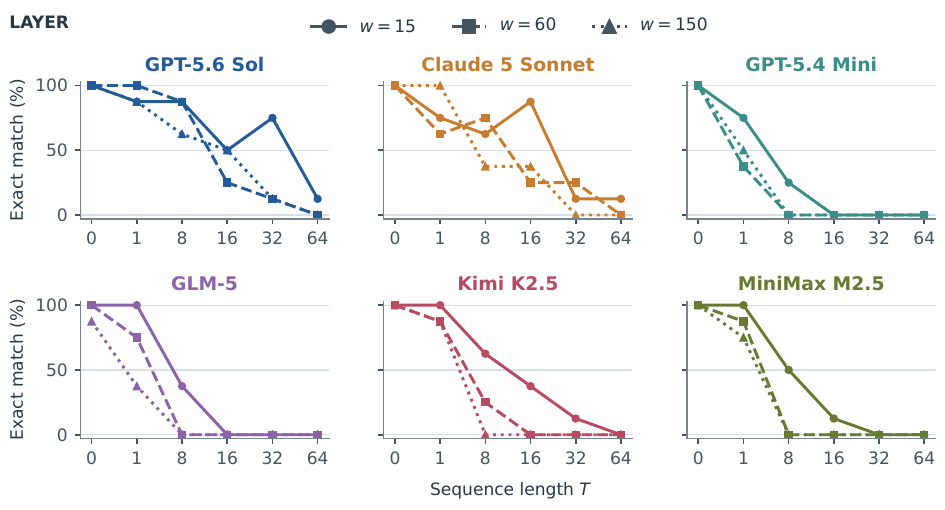}
\caption{\task{LAYER} workload profiles at $d=2$, comparing
component-plus-boundary targets 15, 60, and 150. Each model panel reports
EM over eight trajectories per categorical length, with all outcomes retained.
Markers and line styles identify targets; curves are descriptive.}
\label{fig:workloads-full-3}
\end{figure}

For \task{LAYER}, the pooled advantage of the smallest target does not
hold at every length. Claude 5 Sonnet solves 6/8, 5/8, and 8/8 instances
at $T=1$ for targets 15, 60, and 150, respectively. At $T=16$, these
counts are 7/8, 2/8, and 3/8: the smallest target now has the highest count.

\FloatBarrier
\begin{figure}[H]
\captionsetup{width=0.85\linewidth}
\centering
\includegraphics[trim={4.25bp 4.55bp 3.6bp 7bp},clip,width=0.85\linewidth]{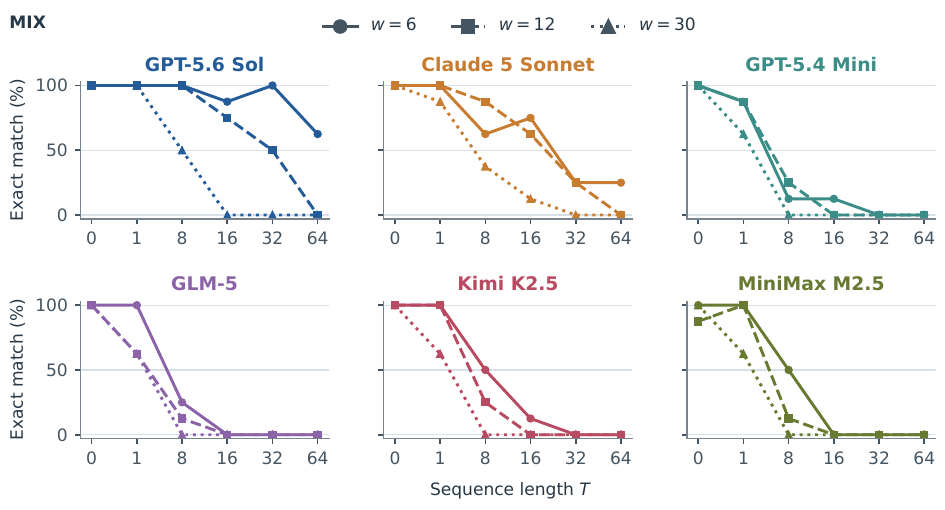}
\caption{\task{MIX} workload profiles at $d=2$, comparing reached-region
targets 6, 12, and 30. Each model panel reports EM over eight trajectories
per categorical length, with all outcomes retained. Markers and line styles
identify targets; curves are descriptive.}
\label{fig:workloads-full-4}
\end{figure}

For \task{MIX}, the smallest target also supports recovery over longer
histories. GPT-5.6 Sol solves all eight one-event instances at every target,
but retains 5/8 successes at $T=64$ only at target 6, compared with zero
at targets 12 and 30. Across the four families, the full curves reveal
where workload contrasts emerge, persist, or reverse as histories lengthen.

Targets use different units across families. Their realized reads and changes
are defined in Appendix~\ref{app:generation} and tabulated in the data package.
For example, over the $T=64$ \task{MIX} histories, increasing the target from
6 to 30 raises mean reference reads per core event from 28.13 to 145.47
and mean changed cells from 3.99 to 20.05. This confirms that the target
change affects both the amount of state consulted and the number of values
altered during reference execution.
The data package also gives exact and valid-state counts at every workload
and length, including the four settings shared with the main experiment.

\FloatBarrier
\subsection{Collection cohorts and paired model comparisons}
\label{app:cohort-results}
\label{app:paired-results}
\begin{table}[!htbp]
\centering
\small
\begingroup
\setlength{\tabcolsep}{4pt}
\renewcommand{\arraystretch}{1.08}
\begin{tabularx}{\columnwidth}{l*{3}{>{\centering\arraybackslash}X}}
\toprule
\tablehead
Model & Original /384 & Additional /384 & Combined /768 \\
\midrule
\rowcolor{TableStripe}
GPT-5.6 Sol & 160 & 154 & 314 \\
\rowcolor{white}
Claude 5 Sonnet & 103 & 98 & 201 \\
\rowcolor{TableStripe}
GPT-5.4 Mini & 13 & 13 & 26 \\
\rowcolor{white}
GLM-5 & 10 & 18 & 28 \\
\rowcolor{TableStripe}
Kimi K2.5 & 23 & 34 & 57 \\
\rowcolor{white}
MiniMax M2.5 & 21 & 27 & 48 \\
\bottomrule
\end{tabularx}
\endgroup
\caption{Exact main-set answers over $T=8,16,32,64$, separated by collection cohort. The original and additional cohorts each contain 96 trajectories, with four prefixes per trajectory. They use disjoint seeds under the same selected generation and evaluation conditions. The combined column is their sum.}
\label{tab:revision-cohorts}
\end{table}

The original and additional collections use disjoint trajectory indices
0--3 and 4--7 under the same generation and evaluation conditions.
Each contributes 744 unique instances per model. Table~\ref{tab:revision-cohorts}
compares their main results over $T=8,16,32,64$, with 384 instances per cohort.
GPT-5.6 Sol and Claude 5 Sonnet have the largest pooled counts in both,
and \task{RELABEL} remains the highest-scoring family for every model in
each cohort. The union is used throughout the main text.

The matched-error illustration also retains its direction in both cohorts.
Kimi K2.5 and MiniMax M2.5 have 42 jointly valid incorrect $T=8$ answers
in each cohort. Their wrong-cell medians are 12.5/85.5 in the original
and 15/57.5 in the addition; the combined 84-instance comparison gives
14/67.

Paired EM comparisons use the identical main instances and the bootstrap
procedure in Appendix~\ref{app:statistics}. GPT-5.6 Sol exceeds Claude 5
Sonnet by 11.46 percentage points at $T=32$ (95\% interval 4.17--18.23)
and 5.73 points at $T=64$ (2.60--8.85). At $T=64$, the other four models
each solve zero or one instance. All fifteen model pairs at both lengths,
including their intervals, are preserved in the supplementary data.

\FloatBarrier
\subsection{Nonmonotonic outcomes on shared trajectories}
\label{app:prefix-patterns}
Table~\ref{tab:revision-prefix-summary} counts histories with an exact answer
at a later prefix after an earlier failure. Every model exhibits this pattern
even when the reconstruction control is excluded.
Including $T=0$ adds one trajectory for GPT-5.4 Mini and GLM-5 and three for
MiniMax M2.5; the other totals are unchanged.

\begin{table}[!htbp]
\centering
\small
\begingroup
\setlength{\tabcolsep}{4pt}
\renewcommand{\arraystretch}{1.08}
\begin{tabularx}{\columnwidth}{l*{2}{>{\centering\arraybackslash}X}}
\toprule
\tablehead
Model & Including $T=0$ & Positive lengths only \\
\midrule
\rowcolor{TableStripe}
GPT-5.6 Sol & 20 & 20 \\
\rowcolor{white}
Claude 5 Sonnet & 51 & 51 \\
\rowcolor{TableStripe}
GPT-5.4 Mini & 13 & 12 \\
\rowcolor{white}
GLM-5 & 5 & 4 \\
\rowcolor{TableStripe}
Kimi K2.5 & 15 & 15 \\
\rowcolor{white}
MiniMax M2.5 & 9 & 6 \\
\bottomrule
\end{tabularx}
\endgroup
\caption{Trajectories with at least one exact later answer after an earlier failed answer, out of 192 per model. The second column includes the reconstruction control; the third uses only positive lengths. Each prefix is a separate request.}
\label{tab:revision-prefix-summary}
\end{table}

These reversals also occur at long histories. Between $T=32$ and $T=64$,
GPT-5.6 Sol loses 34 previously correct instances but gains two; Claude 5
Sonnet loses 27 and gains six. Gains therefore coexist with an aggregate
decline. Each prefix has its own supplied history, reference final state,
and independently queried answer, so the result describes success at
different targets along a shared reference trajectory.

\FloatBarrier

\section{Supplementary State Diagnostics}
\label{app:diagnostics}
Complete-state errors differ in extent, location, and the values that remain
correct. The following analyses extend Section~\ref{sec:state-errors} using
the diagnostic populations defined in Appendix~\ref{app:scoring}.
All-model distributions and matched comparisons are reported separately.

\subsection{Error extent beyond the median}
\label{app:error-extent}
Table~\ref{tab:revision-error-summary} complements the main-text medians
with upper-tail error counts and the frequency of small deviations.
At $T=1$, Claude 5 Sonnet has 44 of 54 valid incorrect states within
five wrong cells, and GPT-5.4 Mini has 75 of 120.
Even when whole-state recovery fails, much of a one-event prediction
can therefore agree with the reference.

\begin{table}[!htbp]
\centering
\small
\begingroup
\setlength{\tabcolsep}{4pt}
\renewcommand{\arraystretch}{1.0}
\begin{tabularx}{\textwidth}{l*{5}{>{\centering\arraybackslash}X}}
\toprule
\tablehead
Model & $T$ & $n$ & Median & P90 & $\leq5$ cells \\
\midrule
\rowcolor{TableStripe}
GPT-5.6 Sol & 1 & 9 & 2.0 & 2.2 & 9 \\
\rowcolor{white}
GPT-5.6 Sol & 16 & 105 & 4.0 & 24.6 & 62 \\
\rowcolor{TableStripe}
GPT-5.6 Sol & 64 & 166 & 33.0 & 123.0 & 12 \\
\rowcolor{white}
Claude 5 Sonnet & 1 & 54 & 2.0 & 8.0 & 44 \\
\rowcolor{TableStripe}
Claude 5 Sonnet & 16 & 111 & 6.0 & 34.0 & 53 \\
\rowcolor{white}
Claude 5 Sonnet & 64 & 44 & 14.0 & 88.9 & 6 \\
\rowcolor{TableStripe}
GPT-5.4 Mini & 1 & 120 & 4.0 & 21.2 & 75 \\
\rowcolor{white}
GPT-5.4 Mini & 16 & 166 & 40.0 & 100.0 & 16 \\
\rowcolor{TableStripe}
GPT-5.4 Mini & 64 & 150 & 114.5 & 183.2 & 3 \\
\rowcolor{white}
GLM-5 & 1 & 95 & 3.0 & 8.6 & 76 \\
\rowcolor{TableStripe}
GLM-5 & 16 & 169 & 41.0 & 99.0 & 15 \\
\rowcolor{white}
GLM-5 & 64 & 148 & 98.5 & 174.6 & 6 \\
\rowcolor{TableStripe}
Kimi K2.5 & 1 & 43 & 2.0 & 9.8 & 32 \\
\rowcolor{white}
Kimi K2.5 & 16 & 153 & 40.0 & 139.2 & 19 \\
\rowcolor{TableStripe}
Kimi K2.5 & 64 & 138 & 107.5 & 184.1 & 4 \\
\rowcolor{white}
MiniMax M2.5 & 1 & 40 & 3.0 & 54.5 & 22 \\
\rowcolor{TableStripe}
MiniMax M2.5 & 16 & 133 & 117.0 & 163.0 & 11 \\
\rowcolor{white}
MiniMax M2.5 & 64 & 136 & 124.5 & 184.0 & 2 \\
\bottomrule
\end{tabularx}
\endgroup
\caption{Error extent at early, intermediate, and long histories, conditional on valid incorrect main answers. Median and P90 count wrong cells in worlds of 252 or 256 cells; P90 uses linear interpolation. The last column counts predictions with at most five wrong cells. Each row has its own subset of size $n$.}
\label{tab:revision-error-summary}
\end{table}

The distributions at longer histories differ in both their center and tail.
For GPT-5.6 Sol, the median rises from four wrong cells at $T=16$ to 33
at $T=64$, while P90 rises from 24.6 to 123.
The fraction within five wrong cells falls from 62/105 to 12/166. Thus,
small deviations become less common as the upper tail extends farther
across the grid.
At $T=64$, GPT-5.4 Mini has median 114.5 and P90 183.2, with only
3/150 states within five cells.

Task conditions further separate error extent within a model.
GPT-5.6 Sol returns 24 valid incorrect states for each of \task{MOVE} and
\task{LAYER} at $T=64$, so both have zero EM, but their wrong-cell medians
are 9.5 and 103. These independently generated task sets illustrate why
complete failure rates can accompany different spatial deviations.
The data package contains all per-instance wrong-cell counts and task-level
quantiles, including lengths not displayed in this table.

\begin{table}[!htbp]
\centering
\small
\begingroup
\setlength{\tabcolsep}{4pt}
\renewcommand{\arraystretch}{1.0}
\begin{tabularx}{\textwidth}{l*{5}{>{\centering\arraybackslash}X}}
\toprule
\tablehead
Model & $T$ & $n_C$ & $C$ acc. (\%) & $n_U$ & $U$ acc. (\%) \\
\midrule
\rowcolor{TableStripe}
GPT-5.6 Sol & 16 & 192 & 95.3 & 192 & 99.0 \\
\rowcolor{white}
GPT-5.6 Sol & 32 & 192 & 88.6 & 192 & 96.1 \\
\rowcolor{TableStripe}
GPT-5.6 Sol & 64 & 191 & 73.6 & 191 & 88.7 \\
\rowcolor{white}
Claude 5 Sonnet & 16 & 172 & 92.7 & 172 & 98.6 \\
\rowcolor{TableStripe}
Claude 5 Sonnet & 32 & 137 & 89.3 & 137 & 96.1 \\
\rowcolor{white}
Claude 5 Sonnet & 64 & 58 & 88.6 & 58 & 91.8 \\
\rowcolor{TableStripe}
GPT-5.4 Mini & 16 & 171 & 68.0 & 171 & 89.4 \\
\rowcolor{white}
GPT-5.4 Mini & 32 & 155 & 51.4 & 155 & 82.6 \\
\rowcolor{TableStripe}
GPT-5.4 Mini & 64 & 151 & 39.6 & 151 & 73.1 \\
\rowcolor{white}
GLM-5 & 16 & 177 & 66.6 & 177 & 88.4 \\
\rowcolor{TableStripe}
GLM-5 & 32 & 170 & 55.3 & 170 & 79.7 \\
\rowcolor{white}
GLM-5 & 64 & 148 & 46.2 & 148 & 76.0 \\
\rowcolor{TableStripe}
Kimi K2.5 & 16 & 171 & 63.9 & 171 & 90.3 \\
\rowcolor{white}
Kimi K2.5 & 32 & 157 & 46.5 & 157 & 83.6 \\
\rowcolor{TableStripe}
Kimi K2.5 & 64 & 138 & 31.8 & 138 & 77.4 \\
\rowcolor{white}
MiniMax M2.5 & 16 & 144 & 46.3 & 144 & 75.4 \\
\rowcolor{TableStripe}
MiniMax M2.5 & 32 & 149 & 38.6 & 149 & 76.3 \\
\rowcolor{white}
MiniMax M2.5 & 64 & 137 & 30.1 & 137 & 73.2 \\
\bottomrule
\end{tabularx}
\endgroup
\caption{Changed- and unchanged-cell accuracy for valid main answers. Each value is a macro-average over the $n_C$ or $n_U$ answers with a nonempty reference subset. $C$ denotes final-reference-changed cells; $U$ denotes final-reference-unchanged cells. Accuracy is undefined for an empty contributing subset. This table covers $T=16,32,64$.}
\label{tab:revision-changed-unchanged-2}
\end{table}

\subsection{Error comparisons on the same instances}
\label{app:matched-valid}
We compare Kimi K2.5 and MiniMax M2.5 at $T=8$ using two common populations.
Their full-set exact counts are 33/192 and 31/192.
Among 131 instances where both return a valid state, their median wrong-cell
counts are four and 39, with mean cell accuracies of 92.8\% and 77.7\%.
This population includes exact predictions.

Restricting to the 84 instances where both predictions are incorrect gives
medians of 14 and 67. Kimi K2.5 has fewer wrong cells on 68 instances,
the same count on three, and more on 13.
The contrast thus appears both when exact states are included and when
only shared failures are compared. It adds information about returned
configurations to their similar full-set EM.

\subsection{Update and preservation accuracy}
\label{app:change-diagnostics}
Figure~\ref{fig:state-errors} classifies valid incorrect predictions by
whether their errors affect changed cells, unchanged cells, or both.
Table~\ref{tab:revision-changed-unchanged-2} complements this classification
with the fraction of each cell group predicted correctly. It averages
per-instance cell accuracies over valid answers at $T=16,32,64$, including
exact states.

At these lengths, every model has lower accuracy on final-changed cells
than on final-unchanged cells. This is compatible with the high prevalence
of the \emph{both} category in Figure~\ref{fig:state-errors}: group accuracy
measures the fraction of values predicted correctly, whereas the partition
records whether any error occurs in each group. Even a prediction with
one wrong changed cell and one wrong unchanged cell belongs to \emph{both},
regardless of how many other cells it gets right. The group accuracies also
include exact answers, while the partition is restricted to incorrect states.
The supplementary data retain both diagnostics at all six lengths.

\subsection{Cell accuracy relative to simple predictions}
\label{app:baseline-accuracy}
\label{app:baselines}
Copy-initial and all-zero obtain zero EM on the 960 positive-length main
instances, yet each matches many individual cells.
Table~\ref{tab:revision-matched-baselines-2} compares cell accuracy at
$T=16,32,64$ on each model's own valid-answer subset.
Within a row, the model and both baselines use exactly the same instances.
The final column adds an instance-level view: how often does the model
improve on copying the supplied world?

\begin{table}[!htbp]
\centering
\small
\begingroup
\setlength{\tabcolsep}{4pt}
\renewcommand{\arraystretch}{1.0}
\begin{tabularx}{\textwidth}{l*{6}{>{\centering\arraybackslash}X}}
\toprule
\tablehead
Model & $T$ & $n$ & Prediction & Copy & Zero & Above copy \\
\midrule
\rowcolor{TableStripe}
GPT-5.6 Sol & 16 & 192 & 98.0 & 63.0 & 48.5 & 192 \\
\rowcolor{white}
GPT-5.6 Sol & 32 & 192 & 93.1 & 56.6 & 48.6 & 191 \\
\rowcolor{TableStripe}
GPT-5.6 Sol & 64 & 191 & 81.6 & 51.0 & 48.8 & 180 \\
\rowcolor{white}
Claude 5 Sonnet & 16 & 172 & 96.8 & 63.5 & 48.9 & 172 \\
\rowcolor{TableStripe}
Claude 5 Sonnet & 32 & 137 & 93.4 & 57.3 & 49.3 & 134 \\
\rowcolor{white}
Claude 5 Sonnet & 64 & 58 & 90.5 & 49.5 & 56.5 & 57 \\
\rowcolor{TableStripe}
GPT-5.4 Mini & 16 & 171 & 82.4 & 63.3 & 48.1 & 160 \\
\rowcolor{white}
GPT-5.4 Mini & 32 & 155 & 69.7 & 56.3 & 48.4 & 129 \\
\rowcolor{TableStripe}
GPT-5.4 Mini & 64 & 151 & 56.1 & 50.6 & 49.8 & 98 \\
\rowcolor{white}
GLM-5 & 16 & 177 & 80.9 & 63.9 & 49.1 & 158 \\
\rowcolor{TableStripe}
GLM-5 & 32 & 170 & 69.6 & 57.9 & 49.6 & 134 \\
\rowcolor{white}
GLM-5 & 64 & 148 & 62.4 & 52.5 & 50.6 & 102 \\
\rowcolor{TableStripe}
Kimi K2.5 & 16 & 171 & 80.6 & 62.9 & 49.1 & 147 \\
\rowcolor{white}
Kimi K2.5 & 32 & 157 & 67.7 & 56.2 & 49.3 & 116 \\
\rowcolor{TableStripe}
Kimi K2.5 & 64 & 138 & 55.6 & 51.1 & 50.2 & 86 \\
\rowcolor{white}
MiniMax M2.5 & 16 & 144 & 64.3 & 62.3 & 50.3 & 76 \\
\rowcolor{TableStripe}
MiniMax M2.5 & 32 & 149 & 58.9 & 55.2 & 49.7 & 76 \\
\rowcolor{white}
MiniMax M2.5 & 64 & 137 & 52.4 & 50.9 & 49.4 & 64 \\
\bottomrule
\end{tabularx}
\endgroup
\caption{Cell accuracy (percent) for model predictions, copy-initial, and all-zero. $n$ is the number of valid main answers; all three predictions in a row use the same instances. Above copy counts model answers with strictly greater cell accuracy than copy-initial. Valid-answer subsets can differ across models. This table covers $T=16,32,64$.}
\label{tab:revision-matched-baselines-2}
\end{table}

For GPT-5.6 Sol, cell accuracy remains above copying on 180/191 valid
$T=64$ predictions. MiniMax M2.5 has mean cell accuracy 52.4\%, compared
with 50.9\% for copying, and improves on copying on 64/137 instances.
In this case, the modest gain in mean accuracy coexists with improvement
on fewer than half of the valid instances. Appendix~\ref{app:matched-valid}
compares models on shared valid instances.

A reference suffix check asks whether executing only late core operations
recovers the target. Suffix-$k$ starts from the initial world, executes every
\task{SET}, and executes core operations only in the final $k$ event
positions. At $T=64$, suffix-8 and suffix-16 recover 0/192 main states,
and suffix-32 recovers 3/192, all in \task{RELABEL}.
These reference checks complement copying with specified partial-execution
baselines. The full reference diagnostics and their definitions are
provided in the supplementary data.

\subsection{Examples of localized and extensive errors}
\label{app:failure-cases}
Three actual valid incorrect predictions illustrate the range of error patterns.
The first, GPT-5.6 Sol on \task{COMPRESS} at $T=16$, has three wrong cells:
two required values are omitted and a third is assigned the wrong nonzero value.
All three belong to the final-changed set.

The second, GPT-5.4 Mini on \task{LAYER} at $T=1$, correctly predicts
all required changes but omits the preserved value at $[2,6]$.
The final GPT-5.4 Mini \task{LAYER} example at $T=64$ differs on
170 of 256 cells. Together, the examples distinguish error extent and
location among predictions that all receive EM zero.

Examples are selected from valid incorrect $d=2$ main responses.
Within each category we take the lexicographically first instance:
GPT-5.6 Sol at $T=16$ with one to five wrong cells; GPT-5.4 Mini at $T=1$
with preservation-only errors; and GPT-5.4 Mini at $T=64$ with at least half
the world wrong.
These are illustrative selections; the aggregate diagnostics report prevalence.

\FloatBarrier
\end{document}